%% file: DualMP.tex
\documentclass[letterpaper]{article} %

\usepackage{aaai2027}

\usepackage[hyphens]{url}  %
\usepackage{graphicx} %
\usepackage{natbib}  %
\usepackage{caption} %
\usepackage{algorithm}
\usepackage{algorithmic}
\usepackage{amssymb}
\usepackage[inline]{enumitem}
\usepackage[all]{foreign}
\usepackage{tabularx}
\usepackage{newfloat}
\usepackage{listings}
\DeclareCaptionStyle{ruled}{labelfont=normalfont,labelsep=colon,strut=off} %
\floatstyle{ruled}
\newfloat{listing}{tb}{lst}{}
\floatname{listing}{Listing}

\usepackage{booktabs}

\usepackage{amsmath}
\usepackage{todonotes}

\newcommand{\mtodoA}[2][]{\todo[#1,backgroundcolor=green!20!white,bordercolor=green!50!black]{\textbf{A:} #2}}

\newif\ifarxiv
\global\arxivtrue

\ifarxiv
\renewcommand{\mtodoA}[2][]{}
\fi

\title{Dual Process Motion Planning}

\author {
    Jiayi Yan\textsuperscript{\rm 1}\ifarxiv\else\equalcontrib\corresponding\fi,
    Francesco Fabiano\textsuperscript{\rm 2}\ifarxiv\else\equalcontrib\fi,
    Alessandro Abate\textsuperscript{\rm 2}
}

\affiliations {
    \textsuperscript{\rm 1}The Chinese University of Hong Kong, Shenzhen\\
    jiayiyan@link.cuhk.edu.cn\\~\\
    \textsuperscript{\rm 2}University of Oxford\\
    francesco.fabiano@cs.ox.ac.uk, alessandro.abate@cs.ox.ac.uk
}

\ifarxiv
    \nocopyright
\fi

\begin{document}

\maketitle

\begin{abstract}
Robotic systems are deeply embedded in both industry and everyday life, where they are expected to act with speed, precision, and reliability.
Classical control and planning methods have long delivered strong guarantees, but often at the cost of computational efficiency and adaptability.
More recently, learning-based approaches have shown promise in overcoming these limitations, enabling agents to leverage experience to accelerate decision-making and address previously intractable problems.
In this work, we bridge these two approaches through a neuro-symbolic perspective on nonlinear motion planning. Inspired by the Thinking Fast and Slow paradigm, we introduce a dual-process architecture that combines the strengths of robust reasoning and learning.
Our framework integrates state-of-the-art symbolic solvers as a ``System-2'' component with experience-driven ``System-1'' modules. A metacognitive controller dynamically orchestrates their interaction, selecting when to rely on fast intuition versus slower, more precise reasoning.
By evaluating the framework across diverse nonlinear benchmark environments, we demonstrate that this architecture yields consistent gains in planning efficiency, accuracy, and generalization, while promoting reuse across tasks.
The results suggest that tightly coupling learning with structured reasoning offers a scalable path toward more capable and adaptive robotic systems.
\end{abstract}

\section{Introduction}

\emph{Motion planning} is a core component of robotic systems and underpins a wide range of applications, including autonomous driving~\citep{DBLP:journals/tiv/PadenCYYF16} and multi-agent coordination~\citep{7810561}.
Its primary objective is to compute a high-quality, collision-free trajectory that connects a start state to a goal state while respecting system dynamics and constraints.
However, real-world environments are often high-dimensional, continuous, and dynamically constrained, which makes motion planning computationally challenging, especially under strict real-time requirements~\citep{DBLP:books/cu/L2006,DBLP:journals/corr/abs-1105-1186}.

Classical motion planning methods can be broadly categorized into experience-based planning using offline trajectory libraries and neural networks, and online optimization-based planning. Experience-based methods, such as those leveraging  neural networks, amortize planning into a learned policy or trajectory generator that maps observations directly to controls or waypoints at runtime, enabling fast online inference and interpolation beyond a finite trajectory library, albeit typically with weaker guarantees~\citep{DBLP:journals/corr/abs-1709-05448}. Their effectiveness depends critically on selecting a trajectory that matches the current scenario, which is difficult in complex environments, hence often leading to suboptimal or unsafe performance. 

In contrast, online optimization-based methods, including \emph{Model Predictive Control} (MPC) ~\citep{MAYNE2000789} and \emph{Control Barrier Function} (CBF)-based safety filters ~\citep{8796030}, compute trajectories by solving constrained optimization problems that explicitly account for system dynamics and environmental constraints.
While these methods reliably provide solutions, they incur in computational overhead, limiting their applicability in time-critical settings.

To bridge this gap, recent work has explored learning-augmented motion planning, where data-driven models are used to improve efficiency and performance~\citep{8463154, article, carvalho2024motionplanningdiffusionlearning}. These methods improve convergence and solution quality, but they typically rely on a fixed pipeline that often hinders their efficiency.
To address this limitation, we draw inspiration from a recent AI paradigm~\citet{DBLP:journals/cacm/FabianoGLMMPRSV25} that is in turn informed by  the dual-system theory~\citet{kahneman2011thinking}. We propose \emph{Dual-MP}, a dual-system motion-planning framework that exploits between fast experience-based and slower online solving.

We instantiate this idea within a SOFAI-style architecture~\citep{pallagani2025sofailab}.
Dual-MP inherits the modular nature of SOFAI: it uses both fast, experience-based and slow, deliberate solvers.
We refer to the former category as System-1 (S1) and to the latter as System-2 (S2).
These are arbitrated through a \textit{metacognitive (MC) agent}.
Dual-MP is equipped with a Neural Network {S1}: a neural policy trained from successful trajectories, and two general {S2} online solvers: one based on MPC and one on CBF. The code used in this work is available
\ifarxiv
 online\footnote{\url{https://github.com/verayannn/System-1-and-System-2-in-Motion-Planning}}.
\else
 in the attached codebase.
\fi

We summarize our main contributions below:

\begin{itemize}
    \item We propose \emph{Dual-MP}, a modular S1/S2 architecture for nonlinear motion planning that arbitrates fast neural planning and deliberate symbolic solving.
    \item We instantiate the framework with a neural S1 policy and two nonlinear S2 solvers, based on MPC and CBFs, under a common MC planner interface.
    \item We add continual learning, where successful trajectories are reused to retrain S1 and improve future fast planning.
    \item We evaluate the framework across diverse nonlinear benchmark families, reporting success, runtime, S1/S2 usage, and trajectory quality.
\end{itemize}

\section{Related work}

\paragraph{Classical Motion Planning.}

Current classical motion planning methods are primarily divided into two popular categories: experience-based planning using neural networks and online optimization-based planning.

Experience-based planning typically uses neural networks to amortize planning from previously solved problems, learning a direct mapping from observations, goals, and local state information to actions, waypoints, or full trajectories~\citep{DBLP:journals/corr/abs-1709-05448,pmlr-v205-fishman23a}. Such methods can provide low-latency inference and strong empirical performance in complex environments by reusing structure learned from expert demonstrations, simulation data, or successful planning experience~\citep{https://doi.org/10.1049/csy2.12020}. While effective, such approaches typically do not preserve explicit models of dynamics, safety constraints, or solver confidence at inference time, which limits their interpretability and their ability to decide when the output should be trusted.

In contrast to experience-based planning methods, online optimization-based methods, including MPC and CBF-based safety filters, compute trajectories or controls by solving constrained optimization problems at runtime~\citep{MAYNE2000789, 7782377}. These methods directly incorporate system dynamics and environmental constraints, yielding high-quality and dynamically feasible solutions. However, they often incur significant computational overhead, which makes real-time deployment challenging in complex or high-dimensional scenarios.

\paragraph{Learning-Augmented Motion Planning.}

To mitigate this, recent work seeks to combine the efficiency of offline planning with the accuracy of online optimization through learning-augmented hybrid approaches.
A common strategy is to use learned models or offline datasets to warm-start online solvers. For example, Memory-of-Motion learns a mapping from task descriptors to state-control trajectories and uses the resulting memory to initialize nonlinear predictive control~\citep{8463154}. \citet{article} use offline motion libraries as reference costs for online MPC, which allows long-horizon offline behaviors to be executed through short-horizon feedback optimization. Transformer-based motion planners learn to restrict or guide the search space from prior data~\citep{johnson2022motionplanningtransformersmotion}, while diffusion-based planners learn multimodal trajectory priors that can be sampled or adapted during planning~\citep{carvalho2024motionplanningdiffusionlearning}. Closely related to our dual-system motivation, \citet{fridovichkeil2018planningfastslowframework} propose a ``Planning, Fast and Slow'' framework in which offline safety computation enables safe switching among online planners. Their method provides a strong safety-aware planning module; in our terminology, such a framework can be viewed as a possible instantiation of \emph{S2}.

Despite these advances, a key limitation of existing hybrid approaches is that they usually invoke the online solver regardless of how well the offline or learned prior matches the current scenario. This leads to unnecessary computation in cases where a retrieved or neural trajectory is already sufficient. More fundamentally, these methods lack a principled mechanism for deciding when online optimization is required. Dual-MP is designed to address exactly this allocation problem.

\section{Background}

\subsection{Nonlinear Model Predictive Control}

Model Predictive Control (MPC) is a widely used optimal control framework for motion planning under dynamic and environmental constraints~\citep{MAYNE2000789}.
At each time step, MPC solves a finite-horizon optimization problem to compute a control sequence that minimizes a cost function while satisfying system dynamics and constraints.

Consider nonlinear dynamics
\(
    \dot{x} = f(x,u),
\)
or, after discretization,
\(
    x_{k+1}=F_d(x_k,u_k),
\)
where \(x_k\in\mathbb{R}^n\) is the state and \(u_k\in\mathbb{R}^m\) is the control. Given the current state \(x_0\) and goal \(x_g\), nonlinear MPC solves
\(
\min_{u_{0:N_h-1}}
\sum_{k=0}^{N_h-1}
\ell(x_k,u_k;x_g)
+
V_f(x_{N_h};x_g)
\)
subject to
\(
x_{k+1}=F_d(x_k,u_k), \qquad
x_k\in\mathcal{X}_{\mathrm{free}}, \qquad
u_k\in\mathcal{U}.
\)
The stage cost is typically chosen as
\(
\ell(x_k,u_k;x_g)
=
\|x_k-x_g\|_{Q}^{2}
+
\|u_k\|_{R}^{2},
\)
where \(Q\succeq 0\) and \(R\succ 0\).
Obstacle avoidance is encoded through nonlinear state constraints defining the collision-free set \(\mathcal{X}_{\mathrm{free}}\)
MPC is accurate because it optimizes over future trajectories, but repeatedly solving the resulting nonlinear program can be computationally expensive.

\subsection{Control Barrier Functions}

Control Barrier Functions (CBFs) provide a complementary approach for enforcing safety constraints in control systems~\citep{8796030}. Unlike MPC, the CBF-QP does not optimize an entire future trajectory; instead, it acts as a safety filter that enforces local obstacle-avoidance constraints.

For a nonlinear control-affine system
\(
    \dot{x}=f(x)+g(x)u,
\)
let the safe set for obstacle \(i\) be
\(
    \mathcal{C}_i=\{x:h_i(x)\geq 0\},
\)
where \(h_i(x)\) is positive outside the obstacle and negative inside it. Forward invariance of \(\mathcal{C}_i\) can be encouraged by enforcing
\(
    L_f h_i(x) + L_g h_i(x)u + \gamma h_i(x) \geq 0,
\)
where \(L_f h_i\) and \(L_g h_i\) are Lie derivatives and \(\gamma>0\) controls how aggressively the controller moves away from the safety boundary.

At each timestep, the CBF controller solves the quadratic program
\(
\min_{u}
\|u-u_{\mathrm{ref}}\|^2
\)
subject to
\(
L_f h_i(x) + L_g h_i(x)u + \gamma h_i(x) \geq 0,
\quad \forall i,
\qquad
u\in\mathcal{U}.
\)
Here, \(u_{\mathrm{ref}}\) is a nominal control input, such as a goal-directed command or a command proposed by the neural System-1 policy.

\subsection{Thinking Fast and Slow in AI}

\citet{kahneman2011thinking} describe human decision making as the interaction between two complementary processes: a fast, intuitive, experience-driven \emph{System-1} and a slower, deliberative, reasoning-based \emph{System-2}.
Recent AI architectures, including SOFAI~\citep{pallagani2025sofailab,DBLP:journals/cacm/FabianoGLMMPRSV25}, adapt this dual-process principle to machine decision-making by combining fast and slow solvers with a metacognitive agent that arbitrates them~\citep{ganapini2021thinkingfastslowai}.
These architectures have proven successful in tackling settings closely related to motion planning, such as classical planning and constrained grid navigation, as shown in~\citet{DBLP:journals/cacm/FabianoGLMMPRSV25}.

Formally, SOFAI defines a decision architecture composed of three components: 
\begin{enumerate*}[label=(\roman*)]
    \item a set of fast {S1} solvers, typically data-driven and experience-based; 
    \item a set of slow {S2} solvers, based on explicit symbolic reasoning; and 
    \item a centralized metacognitive (\emph{MC}) controller.
\end{enumerate*}
Incoming problem instances automatically trigger one or more S1 solvers, which produce candidate solutions together with confidence estimates.
The MC controller then decides whether to accept the S1 proposal or invoke an S2 solver.
This decision is performed in two stages: a lightweight assessment that evaluates whether the expected solution quality satisfies a task-dependent threshold under resource constraints, followed, when necessary, by a cost-benefit comparison between S1 and S2 execution.
S2 reasoning is activated only if its expected improvement compensates for the additional computational cost.
A key aspect of this framework is that S1 behavior is not static.
Through metacognition, solutions produced or validated by S2 can be used to improve the fast solver over time, effectively distilling deliberative reasoning into reactive policies.
In this sense, the architecture supports an iterative refinement process in which expensive symbolic reasoning is gradually amortized into efficient inference, enabling the system to adapt to recurring problem distributions while reducing reliance on S2 computation.

\section{Dual-MP}

As mentioned above, Dual-MP solves motion planning through metacognitive arbitration.
A planning query is defined as
\(
q=(f,\mathcal{O},x_0,x_g,\mathcal{X},\mathcal{U}),
\)
where \(f\) denotes the nonlinear system dynamics, \(\mathcal{O}\) is the obstacle map, \(x_0\) and \(x_g\) are the start and goal states, \(\mathcal{X}\) is the workspace, and \(\mathcal{U}\) is the admissible control set.
In this work, we consider discrete-time nonlinear reach--avoid problems of the form
\(
x_{t+1}=F_d(x_t,u_t),
\)
where \(F_d\) is the discretized nonlinear dynamics used by the planner.\mtodoA{[perhaps discrete-time framework should introduced at the outset, above. ]}

In the following, we define the main components of Dual-MP, namely the S1 and S2 solvers, as well as the functionalities required from the MC module.
In particular, as detailed in Section~\ref{sec:results}, we consider two basic Dual-MP configurations obtained by combining the S1 solver with two S2 solvers.

\subsection{Neural S1}

Our System-1 is implemented as a neural reactive policy. This follows the common approach of imitation learning to predict low-level controls from local environment observations and goal information~\citep{DBLP:journals/corr/abs-1709-05448}.

Given the current rollout context, local obstacle information, nonlinear dynamics features, and the goal direction, the policy predicts a control input directly:
\(
u_t^{\mathrm{S1}}
=
\pi_\theta(c_t,s,d_t,g_t),
\)
where \(c_t\) is a fixed-length window of recent states in a local coordinate frame, \(s\) encodes the local obstacle situation, \(d_t\) encodes the nonlinear dynamics at the current state, and \(g_t\) is the local goal vector.

The policy uses a lightweight convolutional architecture. A one-dimensional convolutional encoder processes the recent state context \(c_t\), while separate multilayer encoders process the situation vector, dynamics features, and goal vector. The resulting features are concatenated and passed through a feedforward control head:
\[
h_t =
[
\phi_c(c_t),
\phi_s(s),
\phi_d(d_t),
\phi_g(g_t)
],
\qquad
u_t^{\mathrm{S1}}=\pi_\theta(h_t).
\]

This design keeps S1 inexpensive at runtime while still conditioning the control prediction on both geometry and nonlinear dynamics.

As mentioned, S1 is trained by imitation from successful System-2 trajectories.
In particular, the bootstrap demonstrations are solved offline by System-2, and the System-2 recoveries are collected online during continual learning.
For each state-control pair, the supervised target is the expert control \(u_t^\star\), with auxiliary single-step losses that encourage accurate one-step motion, correct direction, and progress toward the goal, and a multi-step rollout term that penalizes the drift accumulated when the policy is run in closed loop:

\[
\begin{aligned}
\mathcal{L}(\theta)
={}&
\lambda_u \ell_u
+ \lambda_x \ell_x
+ \lambda_{\mathrm{dir}}\ell_{\mathrm{dir}}
+ \lambda_{\mathrm{spd}}\ell_{\mathrm{spd}} \\
&+
\lambda_{\mathrm{prog}}\ell_{\mathrm{prog}}
+ \lambda_{\mathrm{roll}}\ell_{\mathrm{roll}}
+ \lambda_{\mathrm{smo}}\ell_{\mathrm{smo}}
+ \lambda_{\mathrm{obs}}\ell_{\mathrm{obs}} .
\end{aligned}
\]

Here, \(\ell_u\) is a Huber control-imitation loss, \(\ell_x\) penalizes one-step prediction error under the nonlinear dynamics, \(\ell_{\mathrm{dir}}\) encourages alignment with the expert displacement, \(\ell_{\mathrm{spd}}\) matches step magnitude, and \(\ell_{\mathrm{prog}}\) penalizes insufficient progress toward the goal. The remaining three terms are evaluated on a differentiable \(H\)-step rollout of the policy under the nonlinear dynamics, run once every \(N\) supervised batches (\(H=N=8\)): \(\ell_{\mathrm{roll}}\) penalizes deviation of the rolled-out positions from the expert positions, \(\ell_{\mathrm{smo}}\) penalizes large control changes, and \(\ell_{\mathrm{obs}}\) penalizes proximity to obstacles along the rollout.

At execution time, S1 rolls out the learned policy under the nonlinear dynamics. A lightweight one-step safety filter selects among the predicted control, goal-directed controls, scaled controls, and zero control, keeping only candidates whose next segment is collision-free. The resulting trajectory is then verified for collision freedom and goal reach. In the SOFAI setting, S2 is invoked only when the S1 rollout fails this verification.

\subsection{S2 Solvers: CBF and MPC}

Dual-MP can be instantiated with two state-of-the-art S2 solvers: 

\begin{itemize}
    \item \texttt{safe\_control}: a control barrier function (CBF)-based safety filter ~\citep{kim2025how}
    \item \texttt{acados}: a nonlinear model predictive control (MPC) planner ~\citep{Verschueren2021}
\end{itemize}

Let us note that both solvers are implemented using their standard formulations from the literature.\mtodoA{[cf intro material above - better discuss utilised parameters/choices within those two introduced frameworks]}

\subsection{The Metacognitive Module}
\label{sec:sofai_cl}

Dual-MP follows the SOFAI principle of coordinating fast S1 solvers and slower deliberative S2 solvers through a metacognitive arbitration module~\citep{pallagani2025sofailab}. Depending on the Dual-MP configuration, the S1 and S2 components are instantiated differently; in our experiments, the two configurations are obtained by pairing the S1 solver with two S2 solvers introduced above.

For each new scenario, the metacognitive module first invokes the configured S1 solver. The proposed rollout is accepted only if it passes verification (collision freedom and goal reach - cf. above). If the S1 proposal is rejected, Dual-MP invokes the configured S2 fallback, provided that the available time budget is sufficient relative to the estimated problem difficulty. If S2 does not return a satisfactory solution within the budget, Dual-MP falls back to the best available S1 proposal. Successful S2 rollouts are stored in memory and periodically used to improve S1.

The metacognitive module relies on three quantities to properly arbitrate the two systems: problem difficulty, solution correctness, and solver confidence.
Difficulty is a scenario-level quantity used to characterize the planning instance; correctness measures the quality of a proposed rollout; and confidence estimates whether an S1 proposal is worth verifying.\mtodoA{how is this computed?} Following the SOFAI principle, S2 solvers are treated as deliberative solvers and assigned confidence \(1\), while S1 confidence is computed as described above.\mtodoA{is this the soft correctness below? else, pls provide maths}

\paragraph{Correctness.}
Given a rollout \(\tau=\{x_t\}_{t=0}^T\), correctness is set to \(1\) if the trajectory reaches the goal and remains collision-free.
Otherwise, we assign a soft correctness score based on path length, terminal goal error, and collision penalty:
\(
\mathrm{Correct}(\tau;q)
=
\frac{1}{1+
L(\tau)
+
e_g(\tau)
+
\lambda_{\mathrm{col}}N_{\mathrm{col}}(\tau)},
\)
where
\(
L(\tau)=\sum_{t=0}^{T-1}\|x_{t+1}-x_t\|_1,
\qquad
e_g(\tau)=\|x_T-x_g\|_2.
\)
Here, \(N_{\mathrm{col}}(\tau)\) is a bounded collision penalty and \(\lambda_{\mathrm{col}}\) controls the relative cost of unsafe states. In the experiments reported in this paper, we use a strict correctness threshold of \(1.0\), requiring both goal reaching and collision avoidance. More permissive thresholds are supported by the SOFAI framework and may be useful under strict time constraints.

\paragraph{Difficulty.}
We define the difficulty of a query scenario using simple features to maintain a lightweight process. Let \(\mathcal{O}\) denote the set of rectangular obstacles and let \(\mathcal{W}\) denote the workspace. We first compute the obstacle occupancy
\(
\eta_{\mathrm{occ}}
=
\frac{
\sum_{O_j\in\mathcal{O}}\mathrm{area}(O_j)
}{
\mathrm{area}(\mathcal{W})+\epsilon
}.
\)
Let \(x_{\mathrm{mid}}=(x_0+x_g)/2\) be the midpoint between start and goal, and let
\(
d_{\mathrm{clear}}
=
\min_{O_j\in\mathcal{O}}
\mathrm{dist}(x_{\mathrm{mid}},O_j)
\)
be a clearance estimate from this midpoint to the nearest obstacle. The difficulty is then
\begin{align*}
D(q)
&=
\eta_{\mathrm{occ}}
\left(d_{\mathrm{clear}}+\epsilon\right)^{-1}
\left(1+\log(1+T)\right)
\nonumber\\
&\quad \times
\left(1+\min\left\{10,\left(\epsilon_g+\epsilon\right)^{-1}\right\}\right)
\nonumber\\
&\quad \times
\left(1+\min\left\{10,\left(\epsilon_{\mathrm{safe}}+\epsilon\right)^{-1}\right\}\right), 
\label{eq:difficulty}
\end{align*} 
where \(T\) is the planning horizon, \(\epsilon_g\) is the goal tolerance, and \(\epsilon_{\mathrm{safe}}\) is the collision margin. The first factor captures geometric clutter and clearance, while the remaining factors account for horizon length, goal precision, and safety strictness.

\subsection{SOFAI Continual Learning}

A key advantage of SOFAI-inspired architectures is that S1 solvers can improve over time.
Dual-MP supports this by storing successfully generated S2 trajectories and using them to enrich the corresponding S1 solver.
Specifically, when S2 successfully solves a query, the resulting trajectory is added to the online dataset.
For Neural S1, accumulated S2 trajectories are distilled periodically after a fixed number of episodes by fine-tuning the policy on both the original offline dataset and the newly collected S2 successes.

\section{Experimental Evaluation}
\label{sec:experiments}

All experiments were run on an Intel Xeon Gold 6248 CPU server with 40 cores and 125 GiB of memory, running Ubuntu 22.04.5 LTS. The source code, benchmark data, and additional details, including offline training times, will be provided in the appendix for space reasons.

We evaluate Dual-MP on reach--avoid motion-planning tasks in two-dimensional
continuous environments. Each instance is defined by nonlinear dynamics
\(
    x_{t+1}=F_d(x_t,u_t),
\)
a start state \(x_0\), a goal state \(x_g\), rectangular obstacles \(\mathcal{O}\),
control bounds \(\mathcal{U}\), and workspace bounds \(\mathcal{X}\).
A rollout counts as successful only if it is both collision-free and
goal-reaching; no partial credit is given.
Our experiments address four questions:
\begin{itemize}
    \item{Q1: Can Dual-MP improve state-of-the-art planning?}
    \item{Q2: Does Dual-MP preserve reliability?}
    \item{Q3: How does continual learning affect S1?}
    \item{Q4 Does warm-starting System-2 from the rejected S1 trajectory help System-2 solving?}
\end{itemize}

We evaluate our approach on six benchmark families, each stressing a different aspect of
reach--avoid planning:
\begin{itemize}
    \item \textbf{Large sparse (LS):} a few large obstacles; tests long-horizon
    obstacle avoidance in an open workspace.
    \item \textbf{Dense clutter (DC):} many small rectangles; tests local collision
    avoidance under frequent re-planning pressure.
    \item \textbf{Serial walls (SW):} several wall-with-gap structures in sequence;
    tests repeated constrained passages, where a single missed gap is
    unrecoverable.
    \item \textbf{Maze branching (MB):} intersecting wall structures with multiple
    candidate routes; tests global route commitment.
    \item \textbf{Long slalom (LSM):} a long corridor with alternating offset
    barriers; tests sustained tracking over a horizon far longer than the
    MPC prediction window.
    \item \textbf{Bugtrap (BT):} concave trap layouts; tests robustness to locally
    attractive but globally poor routes.
\end{itemize}

Instances within a family share the workspace scale, the start/goal pair, and
the reach--avoid objective, and vary in obstacle layout and in the parameters
of the nonlinear drift field.
For every family, we generate three disjoint instance sets from separate
seeds: a \emph{training} set of \(100\) instances used only to bootstrap S1, an
\emph{evaluation} set of \(500\) instances used for the results in
Tables~\ref{tab:main_results} and~\ref{tab:per_family}, and a held-out
\emph{probe} set of \(500\) instances used exclusively to measure the effect of
continual learning. The probe set is never trained on and is identical across
all blocks and all methods, so probe curves are directly comparable over time.

All of the generated instances are run on six different solvers/Dual-MP configurations, listed in Table~\ref{tab:planner_configs}.
The three Dual-MP variants differ only in the System-2 they escalate to and in whether
that solver is warm-started from the rejected S1 trajectory.

\begin{table}[t]
\centering
\small
\caption{Planners configuration.}
\label{tab:planner_configs}
\begin{tabular}{lcc}
\toprule
\textbf{Name} & \textbf{System-1} & \textbf{System-2} \\
\midrule
NN                 & Neural policy & --    \\
CBF                & --            & CBF   \\
MPC                & --            & MPC  \\
DMP-CBF         & Neural policy & CBF  \\
DMP-MPC         & Neural policy & MPC  \\
DMP-Warm    & Neural policy & MPC (S1 warm start) \\
\bottomrule
\end{tabular}
\end{table}

The various approaches are then evaluated on three main metrics:
\begin{enumerate*}[label=(\roman*)]
    \item \textit{Success rate}, which is the fraction of instances whose returned rollout is collision-free and reaches the goal. 
    \item \textit{Runtime} which represents the time to find a solution: for Dual-MP it is the sum of the S1 call and, when escalation occurs, the S2 call, so escalated instances are charged for both. We report the mean and the \(90\)th percentile.
    \item \textit{Trajectory Quality} evaluated only on successful rollouts, so that it never trades off
    against the success rate. We use a duration-invariant index that is the
    geometric mean of three sub-scores, each in \((0,1]\) and each normalised
    against a scenario-defined rather than a solver-defined reference:
    \(
        Q \;=\; \bigl(\eta_{\text{path}}\cdot\eta_{\text{smooth}}\cdot\eta_{\text{clear}}\bigr)^{1/3}.
    \)
    \(\eta_{\text{path}} = L_{\text{ref}}/L\) is the standard path-optimality ratio
    of the executed length \(L\) against the shortest collision-free path
    \(L_{\text{ref}}\), obtained by grid search around the obstacles~\cite{6377468}.
    \(\eta_{\text{smooth}} = \mathrm{SPARC}_{\text{min-jerk}}/\mathrm{SPARC}\) is
    the spectral arc length of the speed profile expressed relative to an ideal
    minimum-jerk movement~\cite{Flash1985TheCO}, so a minimum-jerk trajectory scores
    \(1.0\); spectral arc length is used because it is the only smoothness measure
    that is simultaneously valid and reliable~\cite{6104119}.
    \(\eta_{\text{clear}} = \min(d_{\min}/r_{\text{body}},1)\) scores the worst
    obstacle clearance against one body radius, capped so that excess conservatism earns nothing. The geometric mean prevents a near-collision or a chattering control from being compensated elsewhere.
    
    Two properties matter for the comparisons below. First, \(Q\) is invariant to
    how fast the trajectory is traversed, so an aggressive controller cannot buy
    quality by saturating its actuator. Second, because the reference is the scenario's own shortest path,
    \(Q\) is comparable across families of very different difficulty.
\end{enumerate*}

For Dual-MP we additionally report how many successes were produced by S1 alone versus by the S2 fallback.

\subsection{Training Protocol}

\paragraph{Bootstrap.}
For each family and each System-2 solver, we solve the \(100\) training
instances with that solver and retain only the collision-free, goal-reaching
trajectories. The base S1 policy is trained on those demonstrations by
behaviour cloning augmented with a short differentiable rollout loss
(horizon \(8\), applied every \(8\) supervised batches). Crucially, the bootstrap teacher is \emph{matched} to the System-2 the
arm will escalate to: DMP-CBF starts from a CBF-taught policy and DMP-MPC from an MPC-taught one.

\paragraph{Continual learning.}
The \(500\) evaluation instances are shuffled once (fixed seed) and presented
in five sequential blocks of \(100\). After each block, S1 is retrained,
warm-started from the previous block's weights, on the union of the bootstrap
demonstrations and all System-2 recoveries collected so far. Recoveries are
obtained in DAgger fashion~\cite{DBLP:journals/corr/abs-1011-0686}: on instances where S1 was rejected,
we relabel states visited by the S1 rollout with the System-2 solver, so the
policy receives corrective actions on its own state distribution rather than
only on the expert's. A replay fraction of \(0.6\) keeps the fixed bootstrap
demonstrations in the mixture and limits drift toward the increasingly hard,
failure-biased recovery set. The retrained model is evaluated on the frozen probe set before the next block begins.

\subsection{Results}
\label{sec:results}

We now present the experimental results of our paper and use these to answer the four research questions we introduced above.
The main results are presented in Table~\ref{tab:main_results}, which reports results macro-averaged over the six
families for the six configurations, and in Table~\ref{tab:per_family} that breaks the same runs down per family.

\begin{table*}[t]
\centering
\small
\caption{Aggregate results for all methods averaged over the six environment families (\(500\) instances per family).}
\label{tab:main_results}
\begin{tabular}{lcccccc}
\toprule
\textbf{Method} & \textbf{Succ. (\%)} & \textbf{Mean RT (ms)} & \textbf{P90 RT (ms)} & \textbf{Mean \(Q\)} & \textbf{P90 \(Q\)} & \textbf{S1/S2 Solves} \\
\midrule
NN       & 36.4 & 230  & 305  & 0.601 & 0.817 & 182 / 0 \\
CBF      & 82.0 & 657  & 1448 & 0.795 & 0.844 & 0 / 475 \\
MPC      & 84.8 & 5708 & 7092 & 0.844 & 0.875 & 0 / 496 \\
DMP-CBF  & 84.1 & 983  & 2279 & 0.734 & 0.850 & 204 / 217 \\
DMP-MPC  & 85.1 & 4979 & 8086 & 0.780 & 0.877 & 141 / 285 \\
DMP-Warm & 83.6 & 5008 & 8161 & 0.782 & 0.877 & 140 / 278 \\
\bottomrule
\end{tabular}
\end{table*}

\begin{table}[t]
\centering
\small
\caption{Per-family results on instance sets (\(n=500\)). LS: large sparse,
DC: dense clutter, SW: serial walls, MB: maze branching, LSM: long slalom,
BT: bugtrap.}
\label{tab:per_family}
\resizebox{\linewidth}{!}{
\begin{tabular}{lcccccc}
\toprule
\textbf{Method} & \textbf{LS} & \textbf{DC} & \textbf{SW} & \textbf{MB} & \textbf{LSM} & \textbf{BT}\\
\midrule
\multicolumn{7}{l}{\emph{Success rate (\%)}} \\
\\
NN       & 82.8 & 74.4 & 12.0 & 19.2 & 13.0 & 17.0 \\
CBF      & 96.0 & 94.0 & 59.7 & 65.0 & 98.2 & 78.8 \\
MPC      & 98.2 & 94.0 & 77.5 & 69.5 & 76.2 & 93.2 \\
DMP-CBF  & 97.0 & 94.8 & 65.6 & 70.0 & 98.2 & 79.2 \\
DMP-MPC  & 98.2 & 95.2 & 77.8 & 69.6 & 76.6 & 93.4 \\
DMP-Warm & 98.2 & 95.2 & 77.2 & 67.8 & 74.6 & 88.8 \\
\midrule
\multicolumn{7}{l}{\emph{Mean planning runtime (ms)}} \\
\\
NN       & 112 & 135 & 230 & 255 & 460 & 185 \\
CBF      & 121 & 234 & 934 & 961 & 1258 & 437 \\
MPC      & 3439 & 3688 & 4886 & 6140 & 12316 & 3781 \\
DMP-CBF  & 349 & 449 & 1131 & 1212 & 1974 & 783 \\
DMP-MPC  & 1473 & 1670 & 4866 & 5988 & 12490 & 3385 \\
DMP-Warm & 1440 & 1715 & 4909 & 6048 & 12458 & 3476 \\
\midrule
\multicolumn{7}{l}{\emph{Mean trajectory quality \(Q\)}} \\
\\
NN       & 0.781 & 0.750 & 0.580 & 0.484 & 0.425 & 0.586 \\
CBF      & 0.817 & 0.817 & 0.778 & 0.772 & 0.801 & 0.787 \\
MPC      & 0.860 & 0.861 & 0.836 & 0.826 & 0.827 & 0.853 \\
DMP-CBF  & 0.794 & 0.763 & 0.696 & 0.679 & 0.741 & 0.729 \\
DMP-MPC  & 0.761 & 0.716 & 0.793 & 0.802 & 0.817 & 0.792 \\
DMP-Warm & 0.762 & 0.709 & 0.814 & 0.790 & 0.817 & 0.796 \\
\bottomrule
\end{tabular}
}
\end{table}

\paragraph{(Q1) Can Dual-MP improve state-of-the-art planning?}

Yes. Dual-MP improves the MPC baseline by filtering out easy instances with S1 and invoking MPC only when needed, which reduces the number of expensive optimisation calls while improving overall success. This yields the strongest success-runtime trade-off among the compared methods.
The largest gains occur in large sparse and dense clutter environments, where S1 frequently solves instances without escalation. Gains are limited in serial walls, maze branching, and long slalom, where most instances still require MPC. By contrast, DMP-CBF is slower than CBF alone because the cost of an S1 rollout outweighs the savings from avoiding an already inexpensive fallback. Thus, the value of dual-process arbitration is determined primarily by the runtime gap between S1 and S2, rather than by S1 accuracy alone.

\paragraph{(Q2) Does Dual-MP preserve reliability?}

Yes. Across the completed families, both Dual-MP variants match or improve the success rate of their respective S2 fallback (Tables~\ref{tab:main_results} and~\ref{tab:per_family}). This follows from the verification gate: S1 is accepted only when its rollout is collision-free and reaches the goal, while rejected rollouts are delegated to S2. S1 can also solve some instances missed by the corresponding fallback solver, particularly in constrained environments.

The main trade-off is trajectory quality. MPC produces the highest-quality trajectories because it explicitly optimises path efficiency and obstacle clearance, whereas accepted S1 rollouts prioritise fast, feasible completion. Consequently, Dual-MP may return trajectories that are longer or pass closer to obstacles than solutions produced by MPC alone. Dual-MP is therefore most suitable when reducing online planning time is more important than retaining the trajectory quality of every MPC solution.

\paragraph{(Q3) How does continual learning affect S1?}

\begin{figure}[t]
\centering
\includegraphics[width=\linewidth]{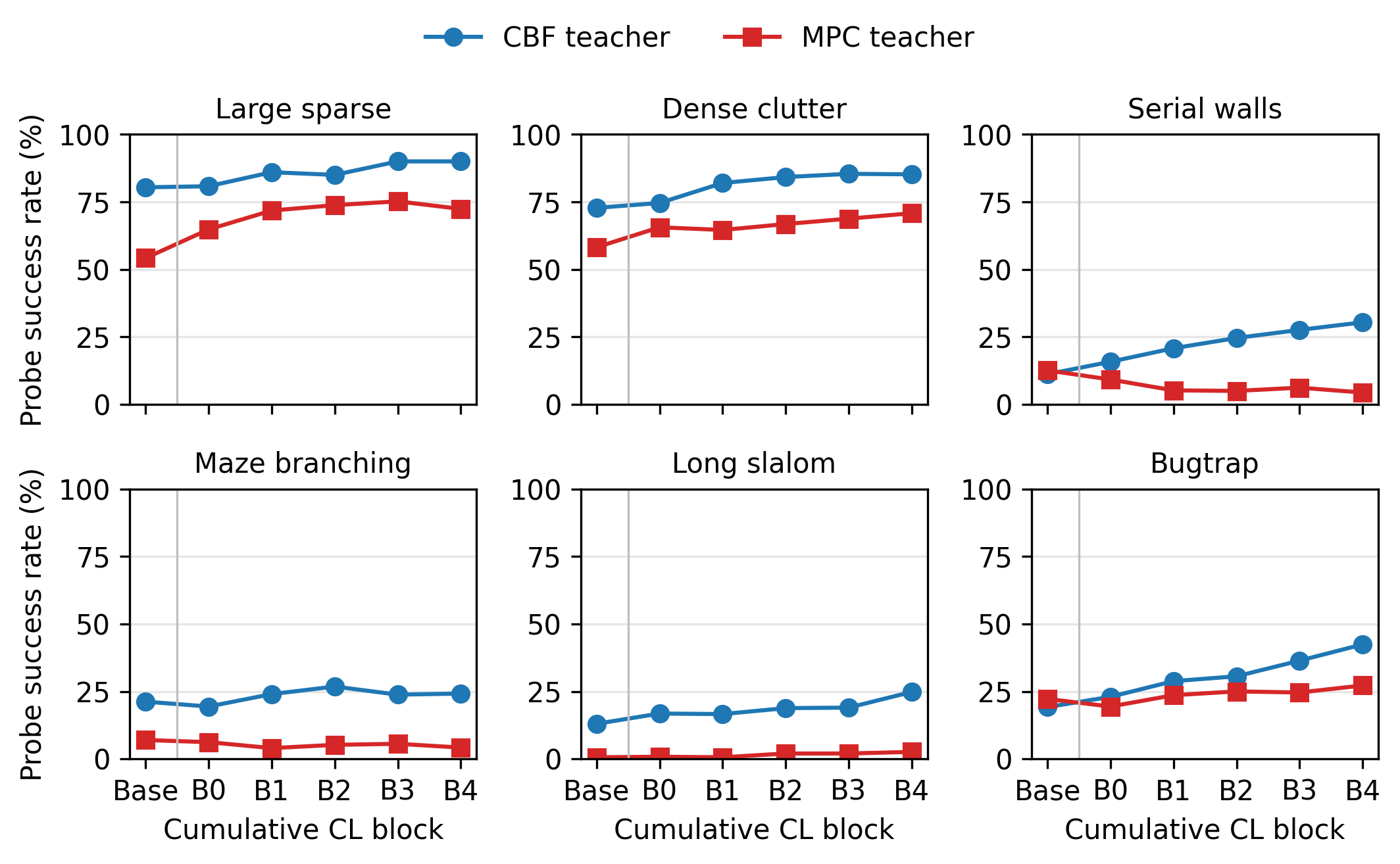}
\caption{Success rate of S1 on the 500-instance probe set as
continual learning accumulates experience. Solid curves show the
CBF- and MPC-taught policies.}
\label{fig:continual_learning_succ}
\end{figure}

\begin{figure}[t]
\centering
\includegraphics[width=\linewidth]{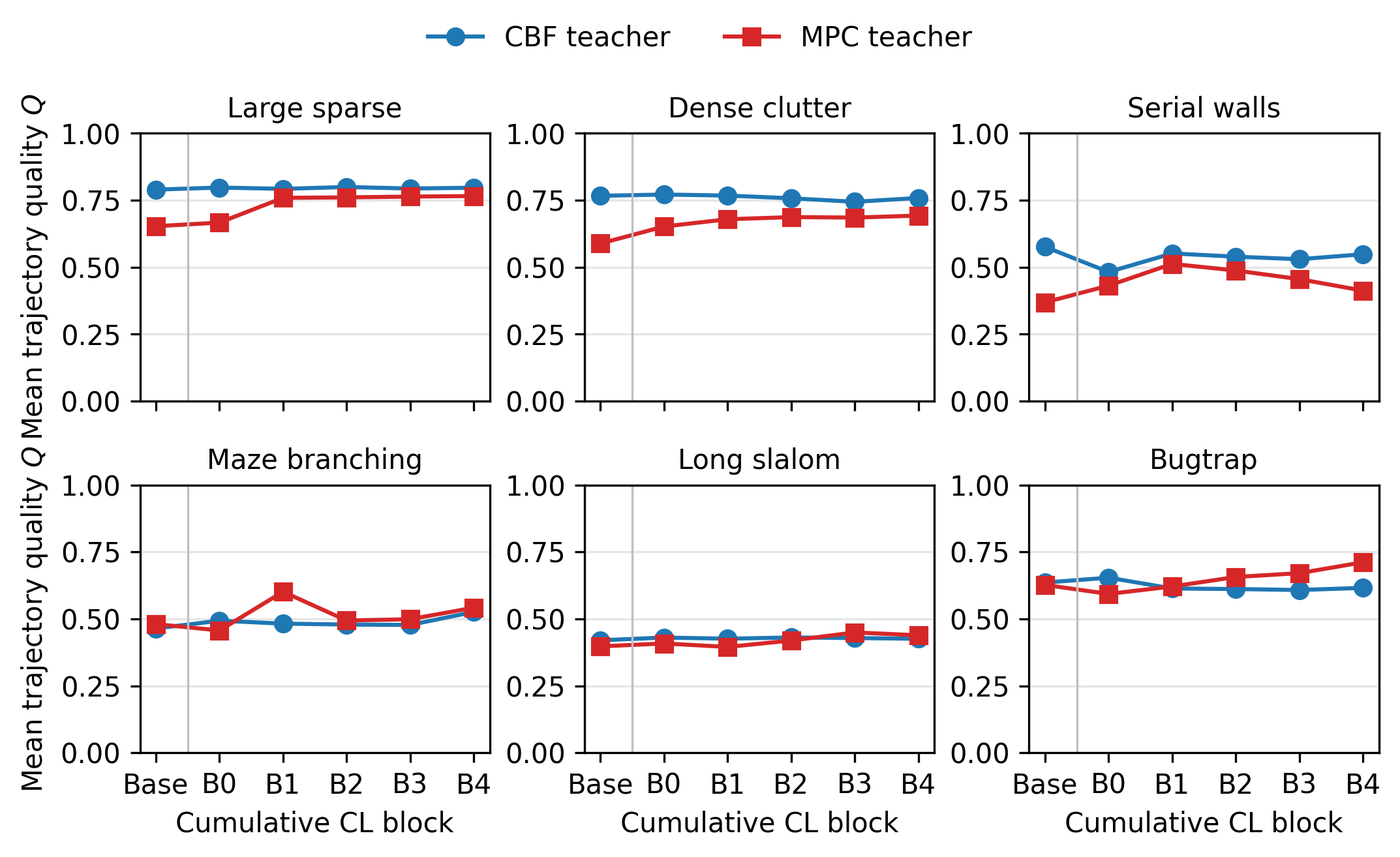}
\caption{Mean quality \(Q\) of successful S1
rollouts on the 500-instance probe set as continual learning
accumulates experience. Quality is evaluated only on successful rollouts.}
\label{fig:continual_learning_quality}
\end{figure}

Table~\ref{tab:cl} reports the cumulative changes in S1 success rate and trajectory quality, while Figures~\ref{fig:continual_learning_succ} and~\ref{fig:continual_learning_quality} show their evolution across training blocks.

Continual learning improves S1, but the type of improvement depends on the S2 teacher. The CBF-taught policy achieves substantial gains in success rate as fallback experience accumulates, increasing the fraction of instances solved directly by S1 and reducing reliance on CBF. The MPC-taught policy primarily improves trajectory quality and increases success in relatively open environments, such as large sparse and dense clutter, but provides limited or inconsistent success gains in more constrained families.

This difference is in the ``teacher'' capabilities: MPC demonstrations often contain long-horizon, globally route-dependent plans that a reactive policy with local observations cannot reliably reproduce after small closed-loop deviations. CBF, by contrast, provides short, local corrective actions that are easier for S1 to imitate.

\begin{table}[t]
\centering
\small
\caption{Continual learning on the held-out (500)-instance probe set. ``Base'' is the bootstrap policy before any online experience. B0--B4 show cumulative performance after each (100)-instance block. \(\boldsymbol{\Delta}\)
is the change from Base to B4, and \(\boldsymbol{\Delta Q}\) is the
corresponding change in mean S1 trajectory quality.}
\label{tab:cl}
\resizebox{\linewidth}{!}{
\begin{tabular}{llccccccrr}
\toprule
\textbf{Family} & \textbf{Arm} & \textbf{Base} & \textbf{B0} & \textbf{B1} & \textbf{B2} & \textbf{B3} & \textbf{B4} & \(\boldsymbol{\Delta}\) & \(\boldsymbol{\Delta Q}\) \\
\midrule
LS  & CBF & 80.4 & 80.8 & 86.0 & 85.0 & 90.0 & 90.0 & $+9.6$  & $+0.007$ \\
    & MPC & 54.2 & 64.8 & 71.8 & 73.8 & 75.2 & 72.4 & $+18.2$ & $+0.113$ \\
DC  & CBF & 72.8 & 74.6 & 82.0 & 84.2 & 85.4 & 85.2 & $+12.4$ & $-0.008$ \\
    & MPC & 58.2 & 65.6 & 64.6 & 66.8 & 68.8 & 70.8 & $+12.6$ & $+0.104$ \\
SW  & CBF & 11.2 & 15.8 & 20.8 & 24.6 & 27.6 & 30.4 & $+19.2$ & $-0.028$ \\
    & MPC & 12.6 & 9.2  & 5.2  & 5.0  & 6.2  & 4.4  & $-8.2$  & $+0.044$ \\
MB  & CBF & 21.2 & 19.4 & 24.0 & 26.8 & 23.8 & 24.2 & $+3.0$  & $+0.062$ \\
    & MPC & 7.0  & 6.2  & 4.0  & 5.2  & 5.6  & 4.2  & $-2.8$  & $+0.061$ \\
LSM & CBF & 13.0 & 16.8 & 16.6 & 18.8 & 19.0 & 24.8 & $+11.8$ & $+0.006$ \\
    & MPC & 0.6  & 0.8  & 0.6  & 2.0  & 2.0  & 2.6  & $+2.0$  & $+0.041$ \\
BT  & CBF & 19.2 & 23.0 & 28.8 & 30.6 & 36.4 & 42.4 & $+23.2$ & $-0.020$ \\
    & MPC & 22.2 & 19.4 & 23.6 & 25.0 & 24.6 & 27.2 & $+5.0$  & $+0.085$ \\
\midrule
\textbf{Mean} & CBF & 36.3 & 38.4 & 43.0 & 45.0 & 47.0 & 49.5 & $+13.2$ & $+0.003$ \\
              & MPC & 25.8 & 27.7 & 28.3 & 29.6 & 30.4 & 30.3 & $+4.5$  & $+0.075$ \\
\bottomrule
\end{tabular}
}
\end{table}

\paragraph{(Q4) Does warm-starting System-2 from the rejected S1 trajectory help System-2 solving?}

Naive warm-starting from a rejected S1 rollout is not useful as shown in Table~\ref{tab:warmstart}. Initialising MPC with the rejected trajectory does not give a consistent runtime benefit, and it can even steer the optimiser toward the same locally poor region that caused S1 to fail. Although warm-starting occasionally helps, the effect is unreliable and can reduce fallback success on the hardest instances. We therefore do not use rejected S1 rollouts as unconditional warm starts. While Dual-MP does compute an S1 confidence score, we did not tune it for warm-start selection because it was not a sufficiently reliable proxy for feasibility, so we omitted that mechanism. We leave this as future work.

\begin{table}[t]
\centering
\small
\caption{Warm-start ablation on instances escalated by both MPC arms. Times are
mean System-2 solver time; ``S2 solves'' counts successful fallbacks.}
\label{tab:warmstart}
\begin{tabular}{lccccc}
\toprule
\textbf{Family} & \textbf{\(n\)} & \textbf{Cold (s)} & \textbf{Warm (s)} & \(\boldsymbol{\Delta}\) & \textbf{S2 solves} \\
\midrule
LS  & 137 & 3.489  & 3.492  & $-0.003$ & 128 / 128 \\
DC  & 150 & 3.764  & 3.771  & $-0.007$ & 126 / 126 \\
SW  & 449 & 4.813  & 4.803  & $+0.010$ & 339 / 337 \\
MB  & 461 & 5.935  & 6.067  & $-0.133$ & 310 / 300 \\
LSM & 477 & 11.096 & 11.057 & $+0.040$ & 370 / 362 \\
BT  & 364 & 3.870  & 4.001  & $-0.131$ & 331 / 311 \\
\bottomrule
\end{tabular}
\end{table}

\subsection{Discussion}

Although Dual-MP is only an initial step toward dual-process motion planning, it delivers promising results; it outperforms the standalone baselines in the reported success and efficiency metrics. This suggests that verified arbitration is not just a conceptual framework, but a practical way to combine learned policies with optimization-based solvers.

The results show that Dual-MP is most effective when the two systems have distinct costs and complementary strengths. A fast S1 policy is most valuable when it avoids expensive MPC calls, but it offers little benefit when the fallback is already cheap, as with CBF. This means that success rate alone is not sufficient to evaluate a hybrid planner: the runtime balance between S1 and S2 largely determines whether arbitration is worthwhile.
Exact rollout verification is central to this behavior. It allows S1 to produce low-latency feasible solutions without letting unsafe or incomplete predictions replace the fallback solver. When S1 fails, S2 takes over, and successful recoveries can then be reused for continual learning. This makes the architecture practical without requiring the neural policy to be perfectly safe in isolation, as long as it is paired with a correct fallback and a verification gate.

Similarly, we note that continual learning is most effective when the teacher’s behavior is compatible with the student’s policy class. CBF provides short, local corrections that improve S1 coverage, whereas MPC often produces longer, route-dependent behaviors that are harder for a reactive policy to reproduce reliably in closed loop. 
This suggests several promising directions for future work: \begin{enumerate*}[label=(\roman*)]
\item training S1 on MPC trajectories and pairing it with CBF,
\item exploring other combinations of fast and slow components, and
\item using a portfolio metacontroller to select among multiple S2 planners rather than relying on a single fallback.
\end{enumerate*}
More generally, future work could optimize different objectives depending on the application, including planning time, trajectory quality, success rate, or even accepting slightly suboptimal solutions when that yields a better overall trade-off.
In the same spirit, better knowledge transfer across environment families could reduce bootstrap cost, either by sharing an S1 policy across tasks or by making continual learning more selective when offline retraining budget is limited. Rejected S1 rollouts should likewise be used more selectively as MPC initializations; a stronger confidence or feasibility estimate could help determine when a proposal is close enough to serve as a useful warm start and when MPC should instead solve the problem from scratch.

Overall, this work provides an initial but promising step toward optimizing not only the individual tools, but also their composition. It shows that the interaction between learned policies, symbolic solvers, verification, and retraining can itself be designed and improved.

\section{Conclusion}

We presented Dual-MP, a dual-process motion-planning architecture that combines a fast neural System-1 policy with a symbolic System-2 fallback. Dual-MP accepts S1 rollouts only when they are collision-free and goal-reaching; otherwise, it delegates planning to either MPC or CBF. Successful S2 recoveries are then reused to improve S1 through continual learning.

Dual-MP with MPC strikes the best balance between success rate and runtime: it achieves the highest success rate while remaining substantially faster than standalone MPC. DMP-CBF is also improved in success rate, but it is slightly slower than CBF because the added S1 rollout overhead outweighs part of the savings. This supports the case for dual-process as it can deliver measurable gains in planning efficiency and performance. At the same time, the results show that continual learning can reduce dependence on System-2. Conversely, naive warm-starting of MPC from rejected S1 trajectories provides no consistent benefit. Overall, Dual-MP offers a promising demonstration that fast-and-slow architecture is a practical mechanism for combining the speed of learned policies with the robustness of optimization-based planning.

\bibliography{aaai2027}

\ifarxiv
\input{Appendix}
\fi

\end{document}

%% file: Appendix.tex
\newpage
\onecolumn
\appendix

\section{Computational Resources and Experimental Configuration}
\label{app:resources}

\subsection{Computational resources}

All experiments were run on an Intel Xeon Gold 6248 CPU server with 40 cores and 125 GiB of memory, running Ubuntu 22.04.5 LTS. 

Overall, generating the six nonlinear benchmark suites, training the System~1 policies, collecting DAgger recovery demonstrations, and evaluating all
System~1/System~2 configurations required several CPU-hours of computation across the experimental batches. A detailed training-time breakdown is provided in Table~\ref{tab:appendix_compute}.

\begin{table}[H]
  \centering
  \small
  \begin{tabularx}{0.95\linewidth}{
    >{\centering\arraybackslash}p{0.23\linewidth}
    |
    >{\centering\arraybackslash}X
    |
    >{\centering\arraybackslash}X
    |
    >{\centering\arraybackslash}X
  }
    \textbf{Environment Family} &
    \textbf{Baseline S1 Training} &
    \textbf{DAgger Collection} &
    \textbf{CL S1 Training} \\
    \hline
    Large sparse (LS)
    & $\sim$6 s
    & 11 s (8--14)
    & $\sim$66 s \\

    Dense clutter (DC)
    & $\sim$84 s
    & 12 s (11--13)
    & $\sim$96 s \\

    Serial walls (SW)
    & $\sim$18 s
    & 32 s (28--35)
    & $\sim$24 s \\

    Maze branching (MB)
    & $\sim$27 s
    & 39 s (37--43)
    & $\sim$18 s \\

    Long slalom (LSM)
    & $\sim$90 s
    & 80 s (75--87)
    & $\sim$102 s \\

    Bugtrap (BT)
    & $\sim$2 s
    & 23 s (21--24)
    & $\sim$36 s \\
  \end{tabularx}
  \caption{Approximate training and dagger collecting time per environment family.}
  \label{tab:appendix_compute}
\end{table}

\subsection{Parameters and Hyperparameters}
\label{app:hyperparameters}

Tables~\ref{tab:appendix_s1} and
\ref{tab:appendix_experiment_settings} summarize the model parameters and the training, benchmark, and reproducibility hyperparameters used in our experiments.

\begin{table}[H]
  \centering
  \small
  \begin{tabularx}{0.82\linewidth}{
    >{\centering\arraybackslash}X
    |
    >{\centering\arraybackslash}p{0.22\linewidth}
  }
    \textbf{Parameter Description} & \textbf{Value} \\
    \hline
    Channel dimensions of the three-layer Conv1d context encoder.
    & $d_{\mathrm{ctx}}\!\rightarrow\!64\!\rightarrow\!128\!\rightarrow\!128$ \\

    Kernel size of the Conv1d context encoder.
    & 3 \\

    Pooling operation applied to the encoded context.
    & Adaptive average pooling \\

    Widths of the situation and dynamics encoders, respectively.
    & 128 / 128 \\

    Width of the goal encoder.
    & 64 \\

    Dimensions of the multilayer perceptron prediction head.
    & $320\!\rightarrow\!256\!\rightarrow\!256\!\rightarrow\!2$ \\

    Hidden-layer width of the neural System~1 policy.
    & 256 \\

    Number of channels in the final convolutional layer.
    & 128 \\

    Dropout probability used throughout the policy.
    & 0.05 \\
  \end{tabularx}
  \caption{Values and descriptions of the neural System~1 model parameters.}
  \label{tab:appendix_s1}
\end{table}

\begin{table}[H]
  \centering
  \small
  \begin{tabularx}{0.82\linewidth}{
    >{\centering\arraybackslash}X
    |
    >{\centering\arraybackslash}p{0.22\linewidth}
  }
    \textbf{Hyperparameter Description} & \textbf{Value} \\
    \hline
    Optimizer used to train the neural System~1 policy.
    & AdamW \\

    Learning rate and weight-decay coefficient.
    & $10^{-4}$ / $10^{-5}$ \\

    Training batch size and validation-data fraction.
    & 64 / 0.10 \\

    Maximum gradient norm used for gradient clipping.
    & 5.0 \\

    Supervised training objective and its transition parameter.
    & Huber loss, $\delta=1.0$ \\

    Number of continual-learning training epochs.
    & 12 \\

    Number of baseline training epochs.
    & 35 \\

    Random seeds used for training, evaluation, and probing.
    & 7 / 8 / 700 \\

    Continual-learning block size and scenario-ordering seed.
    & 100 / 42 \\

    Numbers of evaluation and DAgger workers.
    & 16 / 16 \\

    System~1 integration step and nominal rollout horizon.
    & 0.075 s / 900 steps \\

    Situation-grid resolution and corridor buffer.
    & $25\times25$ / 2 cells \\

    Differentiable rollout horizon and application frequency.
    & 8 steps / every 8 batches \\

    Safety-filter mode used during runtime and differentiable rollouts.
    & Policy \\

    Replay fraction allocated to bootstrap demonstrations.
    & 0.60 \\

    Number of DAgger states sampled from each scenario.
    & 4 \\

    Relative weights assigned to bootstrap and DAgger samples.
    & 1.0 / 1.0 \\

    Planning timeout for each scenario.
    & 60 s \\
  \end{tabularx}
  \caption{Key training, benchmark, and reproducibility hyperparameters.}
  \label{tab:appendix_experiment_settings}
\end{table}

\newpage
\section{Visualising Trajectory Quality}
\label{app:quality-visuals}

\begin{figure}[H]
  \centering
  \begin{minipage}[t]{0.31\linewidth}
    \centering
    \includegraphics[width=\linewidth]{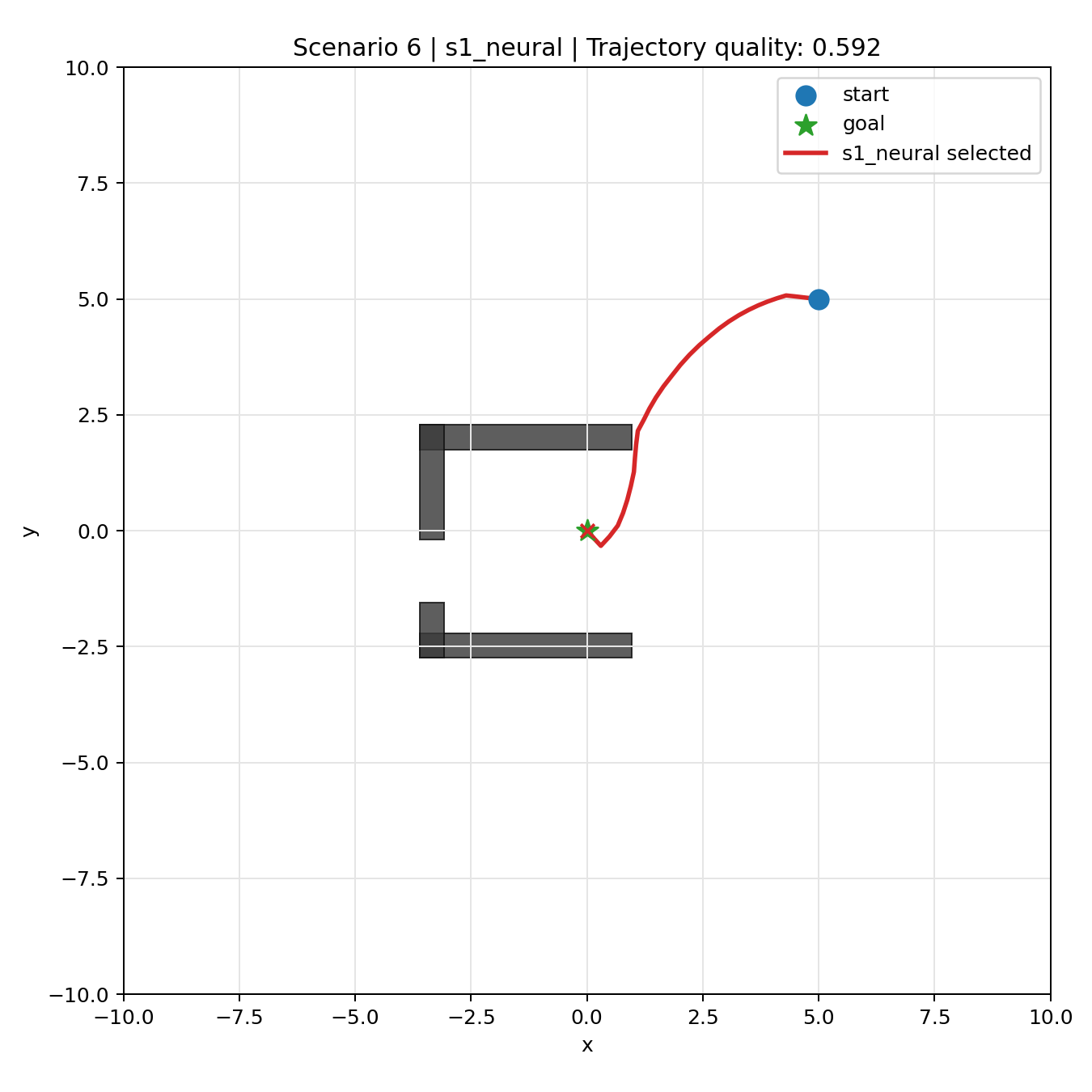}
    \par\smallskip
    \small\textbf{(a) S1-NN.} $Q=0.592$.
  \end{minipage}\hfill
  \begin{minipage}[t]{0.31\linewidth}
    \centering
    \includegraphics[width=\linewidth]{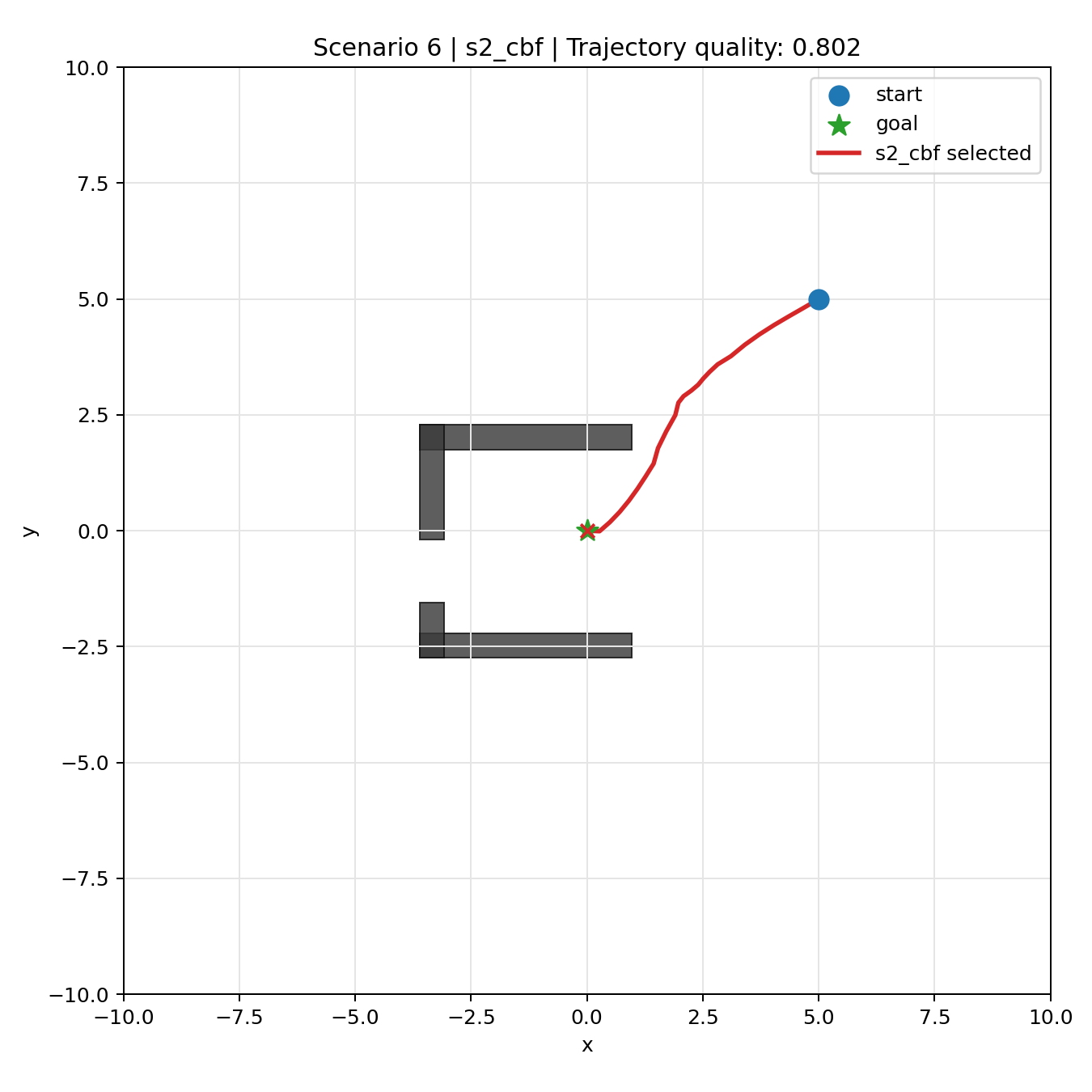}
    \par\smallskip
    \small\textbf{(b) S2-CBF.} $Q=0.802$.
  \end{minipage}\hfill
  \begin{minipage}[t]{0.31\linewidth}
    \centering
    \includegraphics[width=\linewidth]{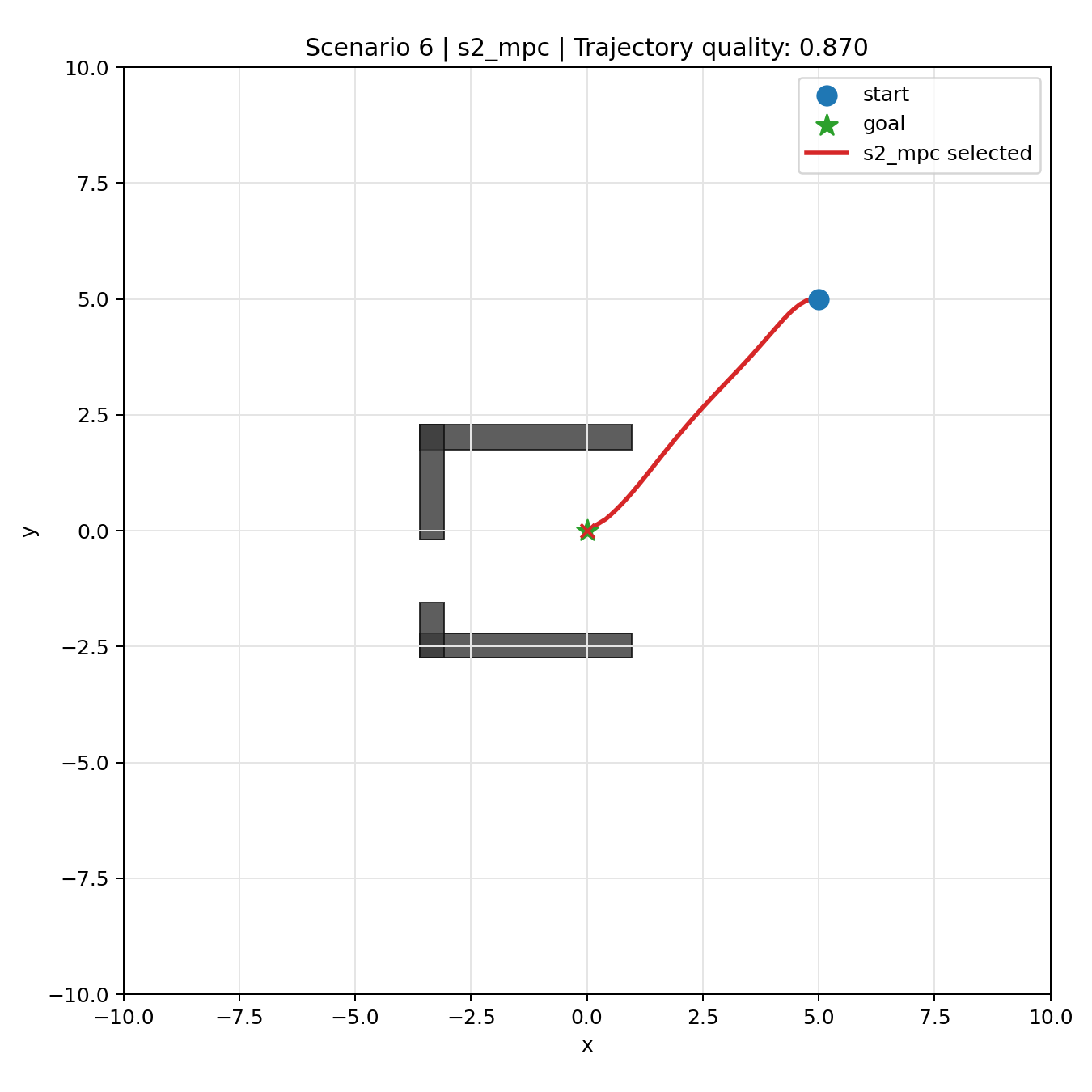}
    \par\smallskip
    \small\textbf{(c) S2-MPC.} $Q=0.870$.
  \end{minipage}
  \caption{Trajectory-quality comparison on Bugtrap scenario~6. Higher $Q$ indicates a more efficient, smoother, and better-cleared trajectory.}
  \label{fig:appendix_quality_visual}
\end{figure}

The trajectory quality is described by a score $Q$ that takes the geometric mean of path efficiency, smoothness, and obstacle-clearance score. A higher score corresponds to a shorter, smoother, and/or better-cleared trajectory. Figure~\ref{fig:appendix_quality_visual} shows
the neural System~1 (NN), System~2 CBF, and System~2 MPC trajectories and their corresponding scores on \emph{the same} scenario. 

\newpage
\section{Visualising the Benchmark Families}
\label{app:domain-visuals}

\begin{figure}[H]
  \centering
  \begin{minipage}[t]{0.31\linewidth}
    \centering
    \includegraphics[width=\linewidth]{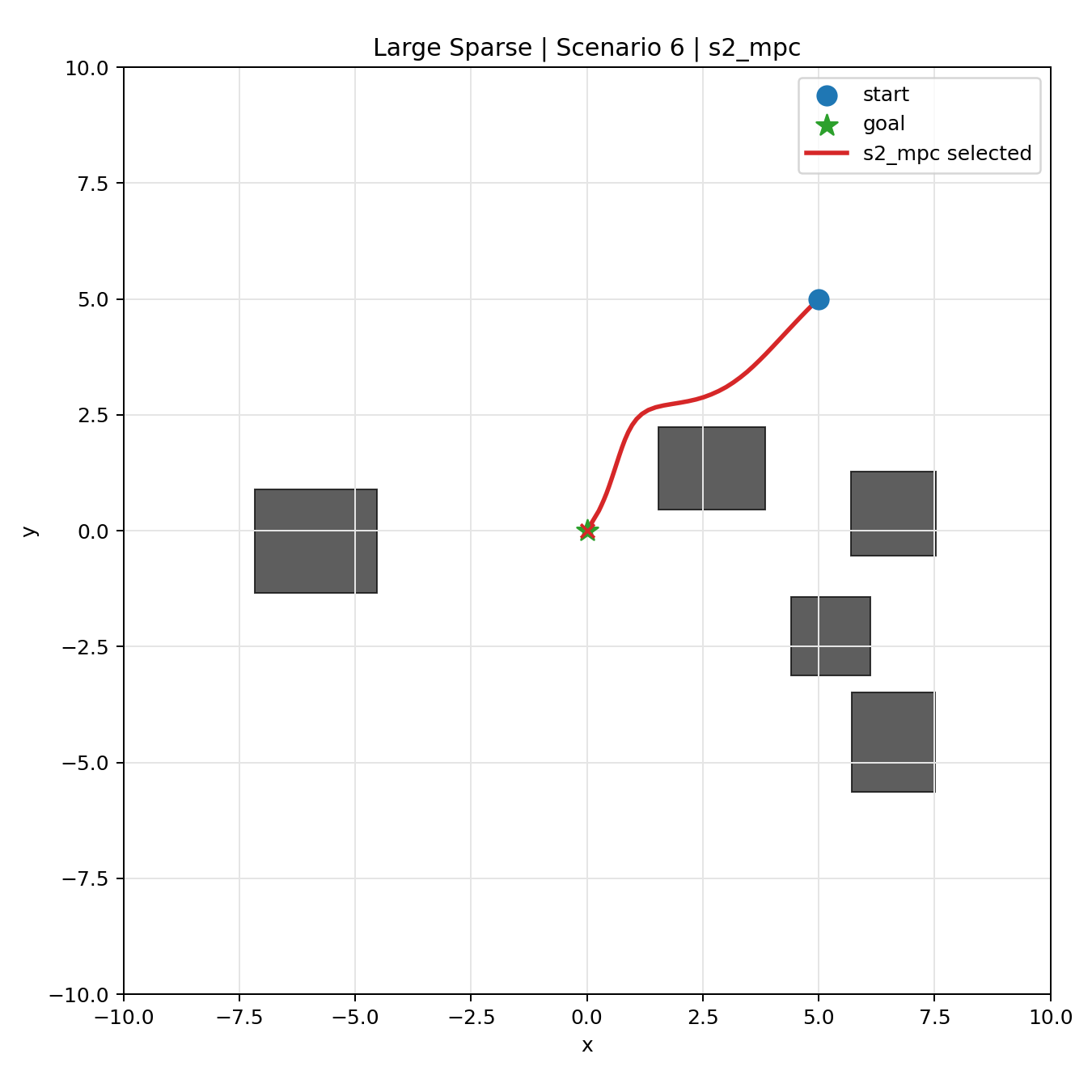}
    \par\smallskip
    \small\textbf{(a) Large Sparse (LS).}
  \end{minipage}\hfill
  \begin{minipage}[t]{0.31\linewidth}
    \centering
    \includegraphics[width=\linewidth]{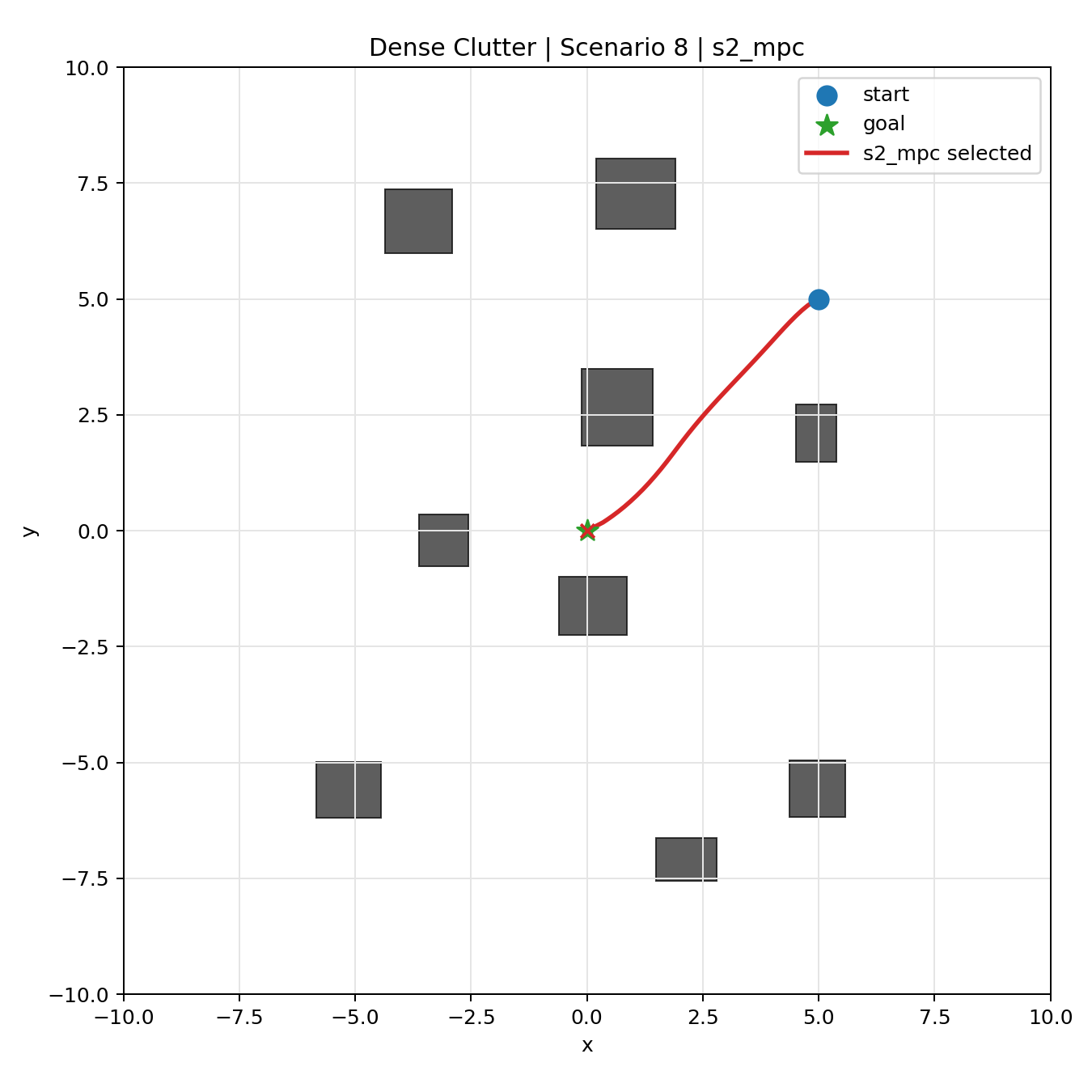}
    \par\smallskip
    \small\textbf{(b) Dense Clutter (DC).}
  \end{minipage}\hfill
  \begin{minipage}[t]{0.31\linewidth}
    \centering
    \includegraphics[width=\linewidth]{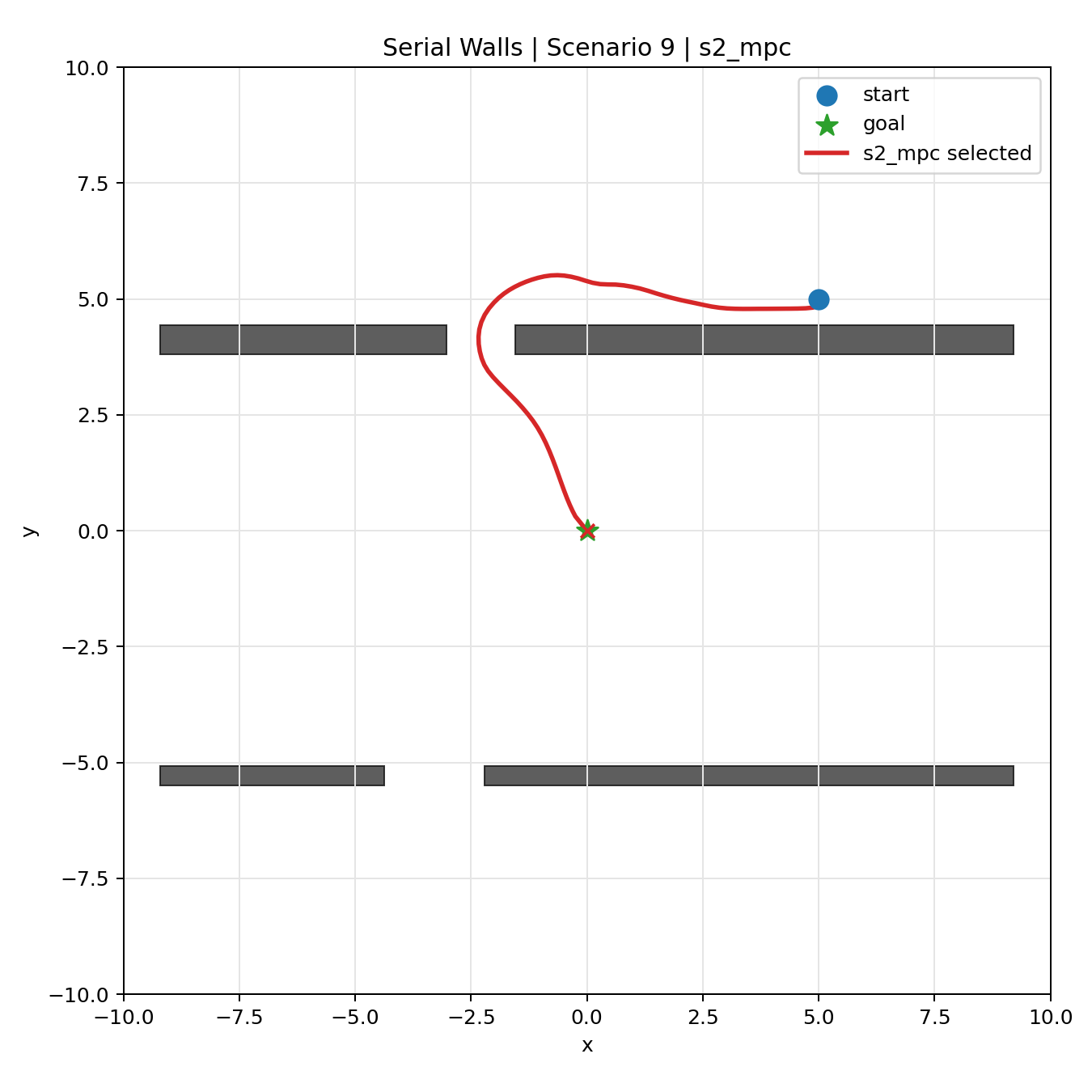}
    \par\smallskip
    \small\textbf{(c) Serial Walls (SW).}
  \end{minipage}

  \vspace{0.75em}

  \begin{minipage}[t]{0.31\linewidth}
    \centering
    \includegraphics[width=\linewidth]{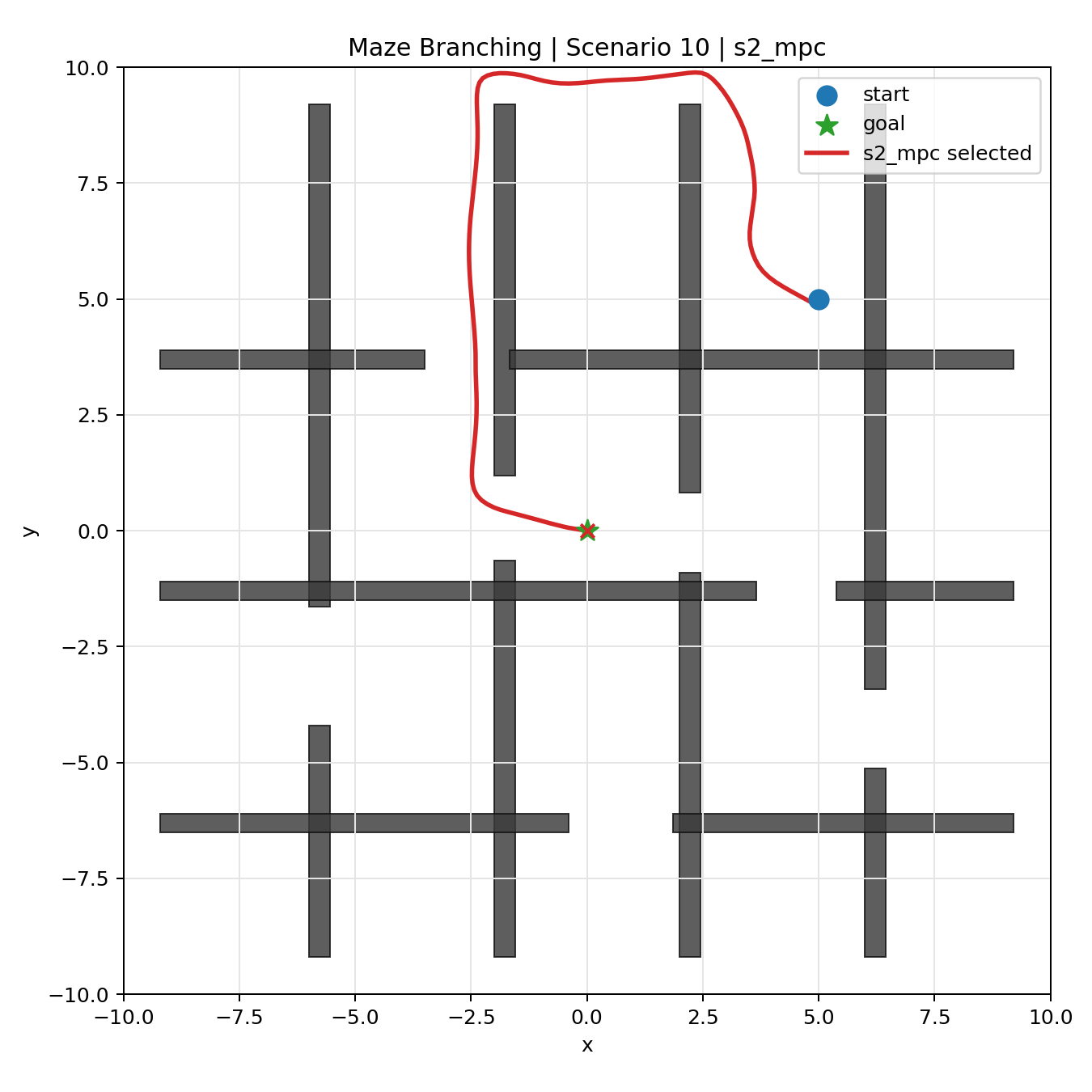}
    \par\smallskip
    \small\textbf{(d) Maze Branching (MB).}
  \end{minipage}\hfill
  \begin{minipage}[t]{0.31\linewidth}
    \centering
    \includegraphics[width=\linewidth]{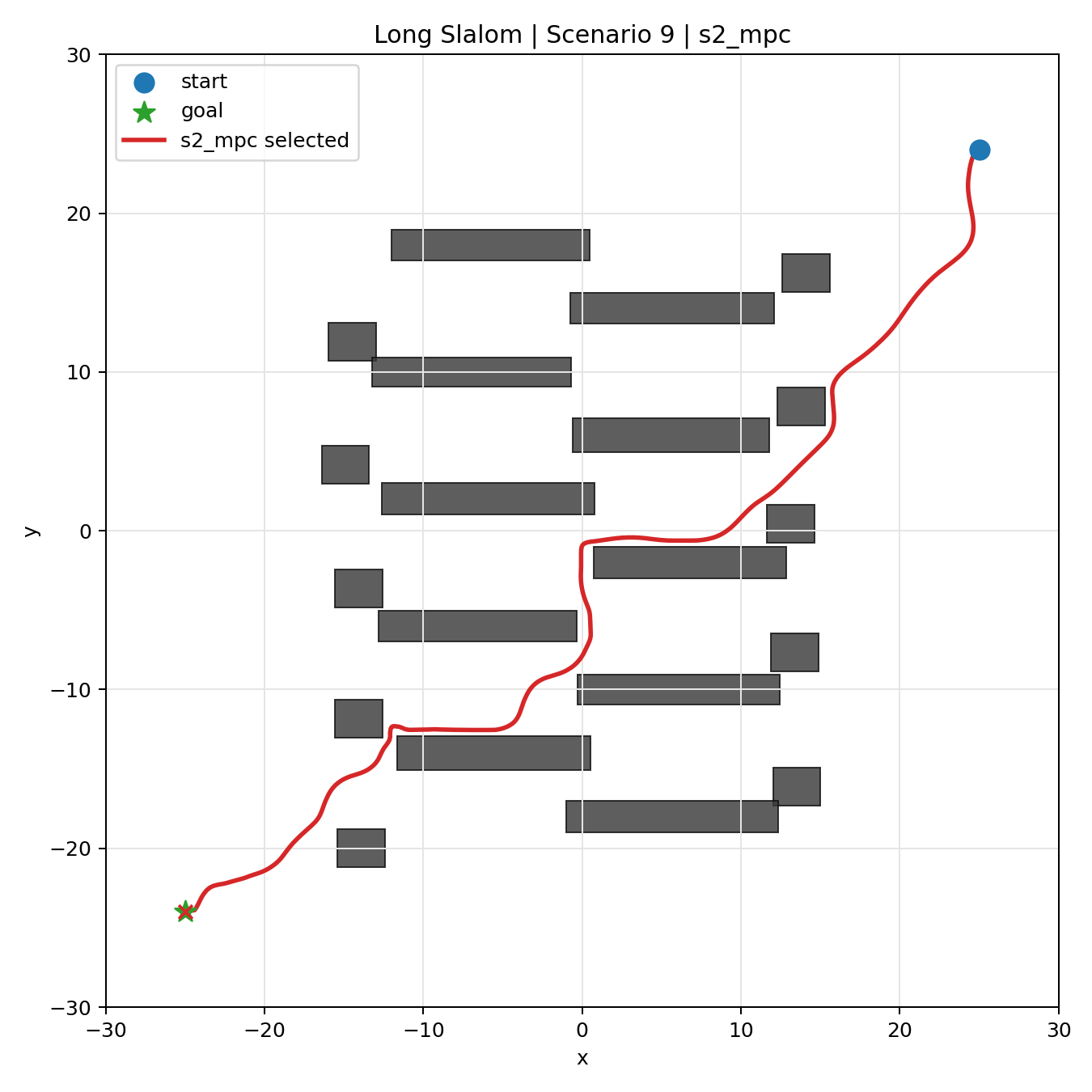}
    \par\smallskip
    \small\textbf{(e) Long Slalom (LSM).}
  \end{minipage}\hfill
  \begin{minipage}[t]{0.31\linewidth}
    \centering
    \includegraphics[width=\linewidth]{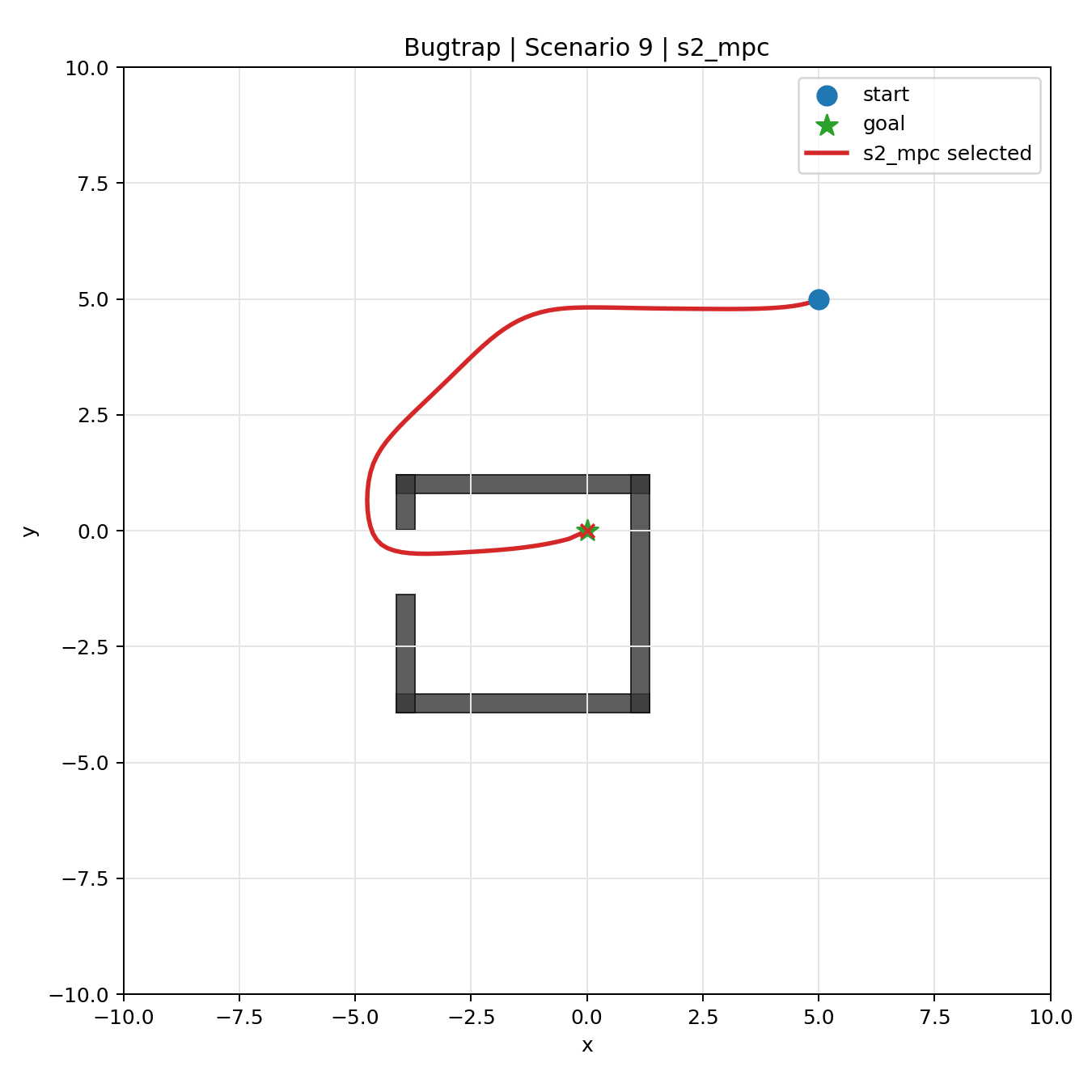}
    \par\smallskip
    \small\textbf{(f) Bugtrap (BT).}
  \end{minipage}
  \caption{Representative nonlinear motion-planning environments used in the
  evaluation. Each panel shows an example map from one benchmark family.}
  \label{fig:appendix_domains}
\end{figure}

Figure~\ref{fig:appendix_domains} provides a visual index of the six nonlinear two-dimensional obstacle-avoidance families and the trajectory computed by System~2 MPC. Large Sparse and Dense Clutter vary obstacle density; Serial Walls and Maze Branching introduce
structured barriers and route choices; Long Slalom creates a long alternating
corridor; and Bugtrap places the robot in a concave trapping geometry.

%% file: aaai2027.bib
@ARTICLE{6377468,
  author={Sucan, Ioan A. and Moll, Mark and Kavraki, Lydia E.},
  journal={IEEE Robotics and Automation Magazine}, 
  title={The Open Motion Planning Library}, 
  year={2012},
  volume={19},
  number={4},
  pages={72-82},
  doi={10.1109/MRA.2012.2205651}}

@ARTICLE{6104119,
  author={Balasubramanian, Sivakumar and Melendez-Calderon, Alejandro and Burdet, Etienne},
  journal={IEEE Transactions on Biomedical Engineering}, 
  title={A Robust and Sensitive Metric for Quantifying Movement Smoothness}, 
  year={2012},
  volume={59},
  number={8},
  pages={2126-2136},
  doi={10.1109/TBME.2011.2179545}}

@inproceedings{Flash1985TheCO,
  title={The coordination of arm movements: an experimentally confirmed mathematical model},
  author={Tamar Flash and Neville Hogan},
  booktitle={Journal of Neuroscience},
  year={1985},
  url={https://api.semanticscholar.org/CorpusID:18355250}
}

@article{DBLP:journals/corr/abs-1011-0686,
  author       = {St{\'{e}}phane Ross and
                  Geoffrey J. Gordon and
                  J. Andrew Bagnell},
  title        = {No-Regret Reductions for Imitation Learning and Structured Prediction},
  journal      = {CoRR},
  volume       = {abs/1011.0686},
  year         = {2010},
  url          = {http://arxiv.org/abs/1011.0686},
  eprinttype   = {arXiv},
  eprint       = {1011.0686},
  bibsource    = {dblp computer science bibliography, https://dblp.org}
}

@inproceedings{kim2025how, 
    author    = {Kim, Taekyung and Beard, Randal W. and Panagou, Dimitra},
    title     = {How to Adapt Control Barrier Functions? A Learning-Based Approach with Applications to a VTOL Quadplane}, 
    booktitle = {IEEE Conference on Decision and Control (CDC)},
    shorttitle = {How to Adapt Control Barrier Functions},
    year      = {2025}
}

@Article{Verschueren2021,
  Title                    = {acados -- a modular open-source framework for fast embedded optimal control},
  Author                   = {Robin Verschueren and Gianluca Frison and Dimitris Kouzoupis and Jonathan Frey and Niels van Duijkeren and Andrea Zanelli and Branimir Novoselnik and Thivaharan Albin and Rien Quirynen and Moritz Diehl},
  Journal                  = {Mathematical Programming Computation},
  Year                     = {2021},
}

@InProceedings{pmlr-v205-fishman23a,
  title = 	 {Motion Policy Networks},
  author =       {Fishman, Adam and Murali, Adithyavairavan and Eppner, Clemens and Peele, Bryan and Boots, Byron and Fox, Dieter},
  booktitle = 	 {Proceedings of The 6th Conference on Robot Learning},
  pages = 	 {967--977},
  year = 	 {2023},
  editor = 	 {Liu, Karen and Kulic, Dana and Ichnowski, Jeff},
  volume = 	 {205},
  series = 	 {Proceedings of Machine Learning Research},
  month = 	 {14--18 Dec},
  publisher =    {PMLR},
  url = 	 {https://proceedings.mlr.press/v205/fishman23a.html}
}

@article{https://doi.org/10.1049/csy2.12020,
author = {Wang, Jiankun and Zhang, Tianyi and Ma, Nachuan and Li, Zhaoting and Ma, Han and Meng, Fei and Meng, Max Q.-H.},
title = {A survey of learning-based robot motion planning},
journal = {IET Cyber-Systems and Robotics},
volume = {3},
number = {4},
pages = {302-314},
doi = {https://doi.org/10.1049/csy2.12020},
url = {https://ietresearch.onlinelibrary.wiley.com/doi/abs/10.1049/csy2.12020},
eprint = {https://ietresearch.onlinelibrary.wiley.com/doi/pdf/10.1049/csy2.12020},
year = {2021}
}

@INPROCEEDINGS{8796030,
  author={Ames, Aaron D. and Coogan, Samuel and Egerstedt, Magnus and Notomista, Gennaro and Sreenath, Koushil and Tabuada, Paulo},
  booktitle={2019 18th European Control Conference (ECC)}, 
  title={Control Barrier Functions: Theory and Applications}, 
  year={2019},
  volume={},
  number={},
  pages={3420-3431},
  doi={10.23919/ECC.2019.8796030}}

@INPROCEEDINGS{7810561,
  author={Hegde, Rashmi and Panagou, Dimitra},
  booktitle={2016 European Control Conference (ECC)}, 
  title={Multi-agent motion planning and coordination in polygonal environments using vector fields and model predictive control}, 
  year={2016},
  volume={},
  number={},
  pages={1856-1861},
  doi={10.1109/ECC.2016.7810561}}

@article{DBLP:journals/tiv/PadenCYYF16,
  author       = {Brian Paden and
                  Michal C{\'{a}}p and
                  Sze Zheng Yong and
                  Dmitry S. Yershov and
                  Emilio Frazzoli},
  title        = {A Survey of Motion Planning and Control Techniques for Self-Driving
                  Urban Vehicles},
  journal      = {{IEEE} Trans. Intell. Veh.},
  volume       = {1},
  number       = {1},
  pages        = {33--55},
  year         = {2016},
  url          = {https://doi.org/10.1109/TIV.2016.2578706},
  doi          = {10.1109/TIV.2016.2578706},
  bibsource    = {dblp computer science bibliography, https://dblp.org}
}

@book{DBLP:books/cu/L2006,
  author       = {Steven M. LaValle},
  title        = {Planning Algorithms},
  publisher    = {Cambridge University Press},
  year         = {2006},
  url          = {http://planning.cs.uiuc.edu/},
  doi          = {10.1017/CBO9780511546877},
  isbn         = {9780511546877},
  bibsource    = {dblp computer science bibliography, https://dblp.org}
}

@article{DBLP:journals/corr/abs-1105-1186,
  author       = {Sertac Karaman and
                  Emilio Frazzoli},
  title        = {Sampling-based Algorithms for Optimal Motion Planning},
  journal      = {CoRR},
  volume       = {abs/1105.1186},
  year         = {2011},
  url          = {http://arxiv.org/abs/1105.1186},
  eprinttype   = {arXiv},
  eprint       = {1105.1186},
  bibsource    = {dblp computer science bibliography, https://dblp.org}
}

@article{MAYNE2000789,
title = {Constrained model predictive control: Stability and optimality},
journal = {Automatica},
volume = {36},
number = {6},
pages = {789-814},
year = {2000},
issn = {0005-1098},
doi = {https://doi.org/10.1016/S0005-1098(99)00214-9},
url = {https://www.sciencedirect.com/science/article/pii/S0005109899002149},
author = {D.Q. Mayne and J.B. Rawlings and C.V. Rao and P.O.M. Scokaert}
}

@article{DBLP:journals/corr/abs-1709-05448,
  author       = {Brian Ichter and
                  James Harrison and
                  Marco Pavone},
  title        = {Learning Sampling Distributions for Robot Motion Planning},
  journal      = {CoRR},
  volume       = {abs/1709.05448},
  year         = {2017},
  url          = {http://arxiv.org/abs/1709.05448},
  eprinttype   = {arXiv},
  eprint       = {1709.05448},
  bibsource    = {dblp computer science bibliography, https://dblp.org}
}

@INPROCEEDINGS{8463154,
  author={Mansard, N. and DelPrete, A. and Geisert, M. and Tonneau, S. and Stasse, O.},
  booktitle={2018 IEEE International Conference on Robotics and Automation (ICRA)}, 
  title={Using a Memory of Motion to Efficiently Warm-Start a Nonlinear Predictive Controller}, 
  year={2018},
  volume={},
  number={},
  pages={2986-2993},
  doi={10.1109/ICRA.2018.8463154}}

@article{article,
author = {Bjelonic, Marko and Grandia, Ruben and Geilinger, Moritz and Harley, Oliver and Medeiros, Vivian and Pajovic, Vuk and Jelavic, Edo and Coros, Stelian and Hutter, Marco},
year = {2022},
month = {06},
pages = {903-924},
title = {Offline motion libraries and online MPC for advanced mobility skills},
volume = {41},
journal = {The International Journal of Robotics Research},
doi = {10.1177/02783649221102473}
}

@misc{carvalho2024motionplanningdiffusionlearning,
      title={Motion Planning Diffusion: Learning and Planning of Robot Motions with Diffusion Models}, 
      author={Joao Carvalho and An T. Le and Mark Baierl and Dorothea Koert and Jan Peters},
      year={2024},
      eprint={2308.01557},
      archivePrefix={arXiv},
      primaryClass={cs.RO},
      url={https://arxiv.org/abs/2308.01557}, 
}

@book{kahneman2011thinking,
  author    = {Kahneman, Daniel},
  title     = {Thinking, Fast and Slow},
  publisher = {Farrar, Straus and Giroux},
  year      = {2011}
}

@article{DBLP:journals/cacm/FabianoGLMMPRSV25,
  author       = {Francesco Fabiano and
                  Marianna Bergamaschi Ganapini and
                  Andrea Loreggia and
                  Nicholas Mattei and
                  Keerthiram Murugesan and
                  Vishal Pallagani and
                  Francesca Rossi and
                  Biplav Srivastava and
                  K. Brent Venable},
  title        = {Thinking Fast and Slow in Human and Machine Intelligence},
  journal      = {Commun. {ACM}},
  volume       = {68},
  number       = {8},
  pages        = {72--79},
  year         = {2025},
  url          = {https://doi.org/10.1145/3715709},
  doi          = {10.1145/3715709},
  bibsource    = {dblp computer science bibliography, https://dblp.org}
}

@misc{ganapini2021thinkingfastslowai,
      title={Thinking Fast and Slow in AI: the Role of Metacognition}, 
      author={Marianna Bergamaschi Ganapini and Murray Campbell and Francesco Fabiano and Lior Horesh and Jon Lenchner and Andrea Loreggia and Nicholas Mattei and Francesca Rossi and Biplav Srivastava and Kristen Brent Venable},
      year={2021},
      eprint={2110.01834},
      archivePrefix={arXiv},
      primaryClass={cs.AI},
      url={https://arxiv.org/abs/2110.01834}, 
}

@ARTICLE{7782377,
  author={Ames, Aaron D. and Xu, Xiangru and Grizzle, Jessy W. and Tabuada, Paulo},
  journal={IEEE Transactions on Automatic Control}, 
  title={Control Barrier Function Based Quadratic Programs for Safety Critical Systems}, 
  year={2017},
  volume={62},
  number={8},
  pages={3861-3876},
  doi={10.1109/TAC.2016.2638961}}

@misc{johnson2022motionplanningtransformersmotion,
      title={Motion Planning Transformers: A Motion Planning Framework for Mobile Robots}, 
      author={Jacob J. Johnson and Uday S. Kalra and Ankit Bhatia and Linjun Li and Ahmed H. Qureshi and Michael C. Yip},
      year={2022},
      eprint={2106.02791},
      archivePrefix={arXiv},
      primaryClass={cs.RO},
      url={https://arxiv.org/abs/2106.02791}, 
}

@inproceedings{pallagani2025sofailab,
  title={SOFAI Lab: A Hands-On Guide to Building Neurosymbolic Systems with Metacognitive Control},
  author={Pallagani, Vishal and Loreggia, Andrea and Fabiano, Francesco and Srivastava, Biplav and Rossi, Francesca and Horesh, Lior},
  booktitle={AAAI Conference on Artificial Intelligence},
  year={2025}
}

@misc{fridovichkeil2018planningfastslowframework,
      title={Planning, Fast and Slow: A Framework for Adaptive Real-Time Safe Trajectory Planning}, 
      author={David Fridovich-Keil and Sylvia L. Herbert and Jaime F. Fisac and Sampada Deglurkar and Claire J. Tomlin},
      year={2018},
      eprint={1710.04731},
      archivePrefix={arXiv},
      primaryClass={cs.SY},
      url={https://arxiv.org/abs/1710.04731}, 
}
